\documentclass[11pt]{article}

\usepackage{jair}
\usepackage{theapa}

\usepackage{amsmath}
\usepackage{amsfonts}
\usepackage{amssymb}
\usepackage{amsthm}
\allowdisplaybreaks[4]

\ifx\pdfoutput\undefined
\else
\fi

\newenvironment{prog}{\begin{array}[t]{@{}l@{}}}{\end{array}}

\newtheoremstyle{theorem}{\topsep}{\topsep}%
     {\itshape}%
     {}%
     {\bfseries}%
     {:}%
     {10pt}%
     {\thmname{#1}\thmnumber{ #2}\thmnote{ (#3)}}%
\theoremstyle{theorem}
\newtheorem{theorem}{Theorem}[section]
\newtheorem{lemma}[theorem]{Lemma}
\newtheorem{corollary}[theorem]{Corollary}
\newtheorem{proposition}[theorem]{Proposition}
\newtheorem{definition}[theorem]{Definition}
\newtheorem{example}[theorem]{Example}

\newcommand{\thm}{\begin{theorem}}
\newcommand{\lem}{\begin{lemma}}
\newcommand{\pro}{\begin{proposition}}
\newcommand{\dfn}{\begin{definition}}
\newcommand{\rem}{\begin{remark} \rm}
\newcommand{\xam}{\begin{example} \rm}
\newcommand{\cor}{\begin{corollary}}
\newcommand{\prf}{\begin{proof}}
\newcommand{\ethm}{\end{theorem}}
\newcommand{\elem}{\end{lemma}}
\newcommand{\epro}{\end{proposition}}
\newcommand{\edfn}{\mbox{}\hfill\qedsymbol\end{definition}}
\newcommand{\erem}{\mbox{}\hfill\qedsymbol\end{remark}}
\newcommand{\exam}{\mbox{}\hfill\qedsymbol\end{example}}
\newcommand{\ecor}{\end{corollary}}
\newcommand{\eprf}{\end{proof}}
\newcommand{\beqn}{\begin{equation}}
\newcommand{\eeqn}{\end{equation}}
\newcommand{\wbox}{\mbox{$\sqcap$\llap{$\sqcup$}}}
\renewcommand{\qedsymbol}{\wbox}

\newenvironment{oldtheorem}[1]
  {\begin{renewcommand}{\thetheorem}{\ref{#1}}}
  {\end{renewcommand}\addtocounter{theorem}{-1}}

\newcommand{\othm}[1]{\begin{oldtheorem}{#1}\begin{theorem}}
\newcommand{\eothm}{\end{theorem}\end{oldtheorem}}
\newcommand{\olem}[1]{\begin{oldthm}{#1}\begin{lemma}}
\newcommand{\eolem}{\end{lemma}\end{oldtheorem}}
\newcommand{\ocor}[1]{\begin{oldtheorem}{#1}\begin{corollary}}
\newcommand{\eocor}{\end{corollary}\end{oldtheorem}}
\newcommand{\opro}[1]{\begin{oldtheorem}{#1}\begin{proposition}}
\newcommand{\eopro}{\end{proposition}\end{oldtheorem}}

\newcommand{\sat}{\models}
\newcommand{\rimp}{\Rightarrow}
\renewcommand{\phi}{\varphi}
\newcommand{\<}{\langle}
\renewcommand{\>}{\rangle}

\newcommand{\riff}{\Leftrightarrow}
\newcommand{\Circ}{\mbox{{\small $\bigcirc$}}}

\newcommand{\abs}[1]{\lvert#1\rvert}
\newcommand{\norm}[1]{\lVert#1\rVert}

\newcommand{\cH}{\mathcal{H}}
\newcommand{\cO}{\mathcal{O}}

\newcommand{\ob}{\mathit{ob}}

\newcommand{\AX}{\textbf{AX}(\Phih,\Phio)}
\newcommand{\AXdyn}{\textbf{AX}_{\mathit{dyn}}(\Phih,\Phio)}
\newcommand{\axiom}[1]{\textbf{#1}}

\newcommand{\intension}[1]{[\![ #1 ]\!]}
\newcommand{\val}[2]{[#1]^{#2}}

\newcommand{\inter}{\ensuremath{\cap}}

\newcommand{\cE}{\ensuremath{\mathcal{E}}}

\newcommand{\Phih}{\Phi_{\mathsf{h}}}
\newcommand{\Phio}{\Phi_{\mathsf{o}}}
\newcommand{\Lh}{\cL_{\mathsf{h}}}

\newcommand{\no}{n_{\mathsf{o}}}
\newcommand{\nh}{n_{\mathsf{h}}}
\newcommand{\Lfoev}{\cL^{\mathit{fo\mbox{-}ev}}}
\newcommand{\Lev}{\cL^{\mathit{ev}}}
\newcommand{\Lw}{\cL^{\mathit{w}}}

\newcommand{\Lfoevdyn}{\cL^{\mathit{fo\mbox{-}ev}}_{\mathit{dyn}}}

\newcommand{\cL}{\ensuremath{\mathcal{L}}}

\renewcommand{\Pr}{\mathrm{Pr}}
\newcommand{\We}{\mathrm{w}}

\newcommand{\truep}{\mbox{\textit{true}}}
\newcommand{\falsep}{\mbox{\textit{false}}}

\newcommand{\phihyp}{\phi_{\scriptscriptstyle\it h}}
\newcommand{\phiobs}{\phi_{\scriptscriptstyle\it o}}
\newcommand{\phiprior}{\phi_{\scriptscriptstyle\it pr}}
\newcommand{\phipost}{\phi_{\scriptscriptstyle\it po}}
\newcommand{\phiprob}{\phi_{\scriptscriptstyle\it p}}
\newcommand{\phiwprob}{\phi_{\scriptscriptstyle\it w,p}}
\newcommand{\phiwup}{\phi_{\scriptscriptstyle\it w,up}}
\newcommand{\phiwf}{\phi_{\scriptscriptstyle\it w,f}}
\newcommand{\phiwcomp}{\phi_{\scriptscriptstyle\it w,c}}

\newcommand{\sequ}[1]{\underline{#1}}

\newenvironment{wideitemize}[1]
   {\begin{list}{$\bullet$}
                     {\setlength{\labelwidth}{#1}
                      \setlength{\leftmargin}{#1}}}
   {\end{list}}
\newenvironment{axiomlist}
   {\begin{wideitemize}{8ex}}
   {\end{wideitemize}}

\newcommand{\bmu}{\boldsymbol{\mu}}
\newcommand{\commentout}[1]{}

\jairheading{26}{2006}{1--34}{11/05}{05/06}
\ShortHeadings{A Logic for Reasoning about Evidence}{Halpern \& Pucella}
\firstpageno{1}

\title{A Logic for Reasoning about Evidence}

\author{\name Joseph Y. Halpern \email halpern@cs.cornell.edu\\
       \addr Cornell University, Ithaca, NY 14853 USA
       \AND
       \name Riccardo Pucella \email riccardo@ccs.neu.edu\\
       \addr Northeastern University, Boston, MA 02115 USA}

\begin{document}

\maketitle

\begin{abstract}
We introduce a logic for reasoning about evidence that essentially
views evidence as a function from prior beliefs (before making an
observation) to posterior beliefs (after making the observation).  
We provide a sound and complete axiomatization for the logic, and
consider the complexity of the decision problem.  Although the reasoning
in the logic is mainly propositional, we allow variables representing
numbers and quantification over them.  This expressive power seems
necessary to capture important properties of evidence.
\end{abstract}

\section{Introduction}

Consider the following situation,
essentially taken from \citeA{HT} and \citeA{FH3}.
A coin is tossed, which is either fair or double-headed.
The coin lands heads. 
How likely is it that the coin 
is double-headed? 
What if the coin is tossed 20 times and it lands heads each time?
Intuitively,
it is much more likely that the coin is double-headed in the
latter case than in the former. 
But how should the likelihood be measured? 
We cannot simply compute
the probability of the coin being double-headed; assigning a
probability to that event requires that we have a prior probability on
the coin being double-headed. 
For example, if the coin was chosen at random from a barrel with one
billion fair coins and one double-headed coin, it is still
overwhelmingly likely that the coin is fair, and that the sequence of 20
heads is just unlucky. 
However, in the problem statement, the prior probability is not
given.
We can show than any given prior probability on the coin 
being double-headed increases significantly as a
result of seeing 20 heads. 
But, intuitively, it seems that we should be able to say 
that seeing 20 heads in a row provides a great deal of \emph{evidence}
in favor of the coin being double-headed without invoking a prior.
There has been a great deal of work in trying to make this intuition
precise, which we now review.  

The main feature of the coin example  is that it involves a combination
of probabilistic outcomes (e.g., the coin tosses) and nonprobabilistic 
outcomes (e.g., the choice of the coin). There has been a great deal
of work on reasoning about systems that combine both probabilistic and
nondeterministic choices; 
see, for example, \citeA{r:vardi85,FZ3,r:halpern88,HT,r:deAlfaro98,r:he97}. 
However, the observations above suggest that if we attempt to formally
analyze
this situation in one of those frameworks, which essentially permit
only the modeling of probabilities, we will not be able to directly
capture this intuition about increasing likelihood.  To see how this
plays out, consider a formal analysis of the situation in the
Halpern-Tuttle \citeyear{HT} framework.  Suppose that Alice
nonprobabilistically chooses one of two coins: a fair coin with
probability $1/2$ of landing heads, or a double-headed coin with
probability $1$ of landing heads. Alice tosses this coin repeatedly.
Let $\phi_k$ be a formula stating: ``the $k$th coin toss lands
heads''.  What is the probability of $\phi_k$ according to Bob, who
does not know which coin Alice chose, or even the probability of
Alice's choice?

According to the Halpern-Tuttle framework, this can be modeled by
considering the set of runs describing the states of the system at each
point in time, and partitioning this set into two subsets, one for
each coin used.  In the set of runs where the fair coin is
used, the probability of $\phi_k$ is $1/2$; in the set of runs where
the double-headed coin is used, the probability of $\phi_k$ is
$1$.  In this setting, the only conclusion that can be drawn is
$(\Pr_B(\phi_k)=1/2) \lor (\Pr_B(\phi_k)=1)$. (This is of course
the probability from Bob's point of view; Alice presumably knows which
coin she is using.)  Intuitively, this seems reasonable: if the fair
coin is chosen, the probability that the $k$th coin toss
lands heads, according to Bob, is $1/2$; if the double-headed coin 
is chosen, the probability is $1$.  Since Bob does not know which of
the coins is being used, that is all that can be said. 

But now suppose that, before the 101st coin toss, Bob learns the
result of the first 100 tosses.  Suppose, moreover, that all of these
landed heads. What is the probability that the 101st coin toss lands
heads?  By the same analysis, it is still either $1/2$ or 
$1$, depending on which coin is used.

This is hardly useful. To make matters worse, no matter how many coin
tosses Bob witnesses, the probability that the next toss lands heads
remains unchanged.  But this answer misses out on some important
information.  The fact that all of  the first 100 coin tosses are
heads
is very strong \emph{evidence} that the coin is in fact 
double-headed.
Indeed,
a straightforward computation using Bayes' Rule shows that if
the prior probability of the coin being double-headed is $\alpha$, then 
after observing that all of the 100 tosses land heads, the probability of the
coin being double-headed becomes
\[ \frac{\alpha}{\alpha+2^{-100}(1-\alpha)} =
\frac{2^{100}\alpha}{2^{100}\alpha+(1-\alpha)}.\] 
However, note that it is not possible to determine the 
posterior
probability that
the coin is double-headed (or that the 101st coin toss is heads) without the
prior probability $\alpha$.    After all, if Alice chooses the
double-headed coin with probability only $10^{-100}$, then it is still
overwhelmingly likely that the coin used is in fact fair, and that Bob
was just very unlucky to see such an unrepresentative sequence
of coin tosses.

None of the frameworks described above for reasoning about
nondeterminism and probability takes the issue of evidence into
account. 
On the other hand, evidence has been discussed extensively in the
philosophical literature.  Much of this discussion occurs in the
philosophy of science, specifically \emph{confirmation theory}, where
the concern has been historically to assess the support that evidence
obtained through experimentation lends to various scientific theories 
\cite{r:carnap62,r:popper59,r:good50,r:milne96}.  
(Kyburg \citeyear{r:kyburg83} provides a good overview of the literature.)

In this paper, we introduce a logic for reasoning about evidence. Our
logic extends 
a
logic defined by Fagin, Halpern and Megiddo
\citeyear{r:fagin90} (FHM from now on) for reasoning about likelihood
expressed as either probability or belief.
The logic has first-order quantification over the reals (so includes
the theory of real closed fields), as does the FHM logic, for reasons
that will shortly become clear.
We add observations to the states, and provide an additional operator
to talk about the evidence provided by particular observations.  We also
refine the language to talk about both the prior probability of
hypotheses and the posterior probability of hypotheses, taking into
account the observation at the states.
This lets us write formulas that talk about the
relationship between the prior probabilities, the posterior
probabilities, and the 
evidence 
provided by 
the observations.

We then provide a sound and complete axiomatization for the logic. To
obtain such an axiomatization, we
seem to 
need first-order quantification in a
fundamental way. Roughly speaking, this is because ensuring that the
evidence operator has the appropriate properties requires us to assert
the existence of suitable probability measures.
It does not seem possible to do this without 
existential quantification. Finally, we 
consider the
complexity of the satisfiability problem.
The complexity problem for the full language requires exponential
space, since it incorporates the theory of real closed fields, for
which an exponential-space lower bound is known \cite{BKR}. 
However, we show that the satisfiability problem for a propositional
fragment of the language, which is still strong enough to allow us to
express many properties of interest, is decidable in polynomial
space.

It is reasonable to ask at this point why we should bother with a logic
of evidence.  Our claim is that many decisions in practical
applications are made on the basis of evidence.  
To take an example from security, consider an enforcement mechanism
used to detect and react to intrusions in a computer system. Such an
enforcement mechanism analyzes the behavior of users and attempts to
recognize intruders. Clearly the mechanism wants to make sensible
decisions based on observations of user behaviors.  How should it do
this? One way is to think of an enforcement mechanism as accumulating
evidence for or against the hypothesis that the user is an
intruder. 
The accumulated evidence can then be used as the basis for a decision to
quarantine a user.  In this context, it is not clear that there is a
reasonable way to assign a prior probability on whether a user is an
intruder.
If we want to specify the behavior of such systems and prove that they
meet their specifications, it is helpful to have a logic that allows
us to do this.  We believe that the logic we propose here is the first
to do so.

The rest of the  paper is organized as follows. 
In the next section, we formalize
a notion of evidence that captures the intuitions outlined above. In
Section~\ref{s:logic}, we introduce our logic for reasoning about
evidence. In Section~\ref{s:axiomatization}, we present an
axiomatization for the logic and show that it is sound and complete with
respect to the intended models.
In Section~\ref{s:decision}, we
discuss the complexity of the decision problem of our logic.  
In Section~\ref{s:normalized}, we examine some alternatives to the
definition of weight of evidence we use.  
For ease of exposition, in most of the paper, we consider a system
where there are only two time points: before and after 
the observation.
In Section~\ref{s:dynamic}, we extend our work to 
dynamic
systems, where there can be multiple pieces of evidence, 
obtained at different points in time.
The proofs of our technical results can be found in the appendix.

\section{Measures of Confirmation and Evidence}\label{s:evidence}

In order to develop a logic for reasoning about evidence, we need to first
formalize an appropriate notion of evidence. 
In this section, we review  various formalizations from the literature,
and discuss the formalization we use. 
Evidence has been studied in depth 
in the philosophical literature,
under the name of \emph{confirmation theory}.  Confirmation theory
aims at determining and measuring the support a piece of evidence
provides an hypothesis. 
As we mentioned in the introduction, many different 
measures of confirmation 
have been proposed in the literature. Typically, a proposal has been
judged on the degree 
to which it satisfies various properties that are considered appropriate
for confirmation. For
example, 
it may be required that
a piece of evidence $e$ confirms an
hypothesis $h$ if and only if $e$ makes $h$ more probable.
We have no desire to
enter the debate as to which class of measures of confirmation is more
appropriate. For our purposes, most confirmation functions are
inappropriate:
they assume that we are given a prior on the set of hypotheses and
observations.  By marginalization, we also have a prior on
hypotheses, which is 
exactly the information we do not have and do not want to assume. 
One exception is measures of evidence that use the 
\emph{log-likelihood ratio}.
In this case, rather than
having a prior on hypotheses and observations, it suffices that 
there be a probability $\mu_h$ on observations for each hypothesis
$h$: intuitively, $\mu_h(\ob)$ is the probability of observing $\ob$ 
when $h$ holds.
Given an observation $\ob$, the degree of confirmation that it provides
for an hypothesis $h$ is
\[l(\ob,h) =
\log\left(\frac{\mu_h(\ob)}{\mu_{\overline{h}}(\ob)}\right),\]  
where 
$\overline{h}$ represents the hypothesis other than $h$ (recall that
this approach applies only if there are two hypotheses).  
Thus, the degree of confirmation is the ratio between these two
probabilities. 
The use of the logarithm is not critical here.  Using it ensures that
the likelihood is positive 
if and only if
the observation confirms the hypothesis.
This approach has been advocated by Good
\citeyear{r:good50,r:good60}, among others.%
\footnote{Another related approach, the \emph{Bayes factor} approach,  is
based on taking the ratio of  \emph{odds} rather than likelihoods
\cite{r:good50,r:jeffrey92}. We remark that in the literature,
confirmation is usually taken with respect to some background
knowledge. For ease of exposition, we ignore background knowledge
here, although it can easily be incorporated into the framework we
present.}

One problem with the log-likelihood ratio measure $l$ as we have 
defined it is that it can be used only to reason about
evidence discriminating between two competing hypotheses, namely
between an hypothesis $h$ holding and the hypothesis $h$ not
holding. We would like a measure of confirmation 
along the lines of the log-likelihood ratio measure, but
that can handle multiple competing hypotheses. There have been a number
of such generalizations, for example, by \citeA{r:pearl88}
and \citeA{r:chan05}. 
We focus
here on the  generalization given by Shafer \citeyear{Shafer82} in the
context of 
the Dempster-Shafer theory of evidence based on belief functions
\cite{r:shafer76}; it was further 
studied by Walley \citeyear{Walley87}. 
The description here is taken mostly from \citeA{HF2}.
While this measure of confirmation has a number of nice properties of
which we take advantage, 
much of the work presented in this paper can be adapted to different
measures of confirmation. 

We start with a finite set $\cH$ of mutually
exclusive and exhaustive hypotheses; thus, exactly one hypothesis holds
at any 
given time. Let $\cO$ be the set of possible observations (or pieces of
evidence).  For simplicity, we assume that $\cO$ is finite.  
Just as in the case of log-likelihood, we also assume that, for each
hypotheses $h\in\cH$, there is a probability measure $\mu_h$ on $\cO$
such that $\mu_h(\ob)$ is the probability of $\ob$ if hypothesis $h$ holds.
Furthermore, we assume that the observations in $\cO$ are relevant to
the hypotheses: for every observation $\ob\in\cO$, there must be an
hypothesis $h$ such that $\mu_h(\ob)>0$. 
(The measures $\mu_h$ are often called \emph{likelihood functions} in the
literature.) We define an \emph{evidence space} (over $\cH$ and $\cO$)
to be a tuple $\cE=(\cH,\cO,\bmu)$, where $\bmu$ is a function that
assigns to every hypothesis $h\in\cH$ the likelihood function
$\bmu(h)=\mu_h$. (For simplicity, we usually write $\mu_h$ for
$\bmu(h)$, when the the function $\bmu$ is clear from context.)
Given an evidence space $\cE$, we define the weight that the
observation $\ob$ lends to hypothesis $h$, written $w_{\cE}(\ob,h)$,
as
\begin{equation}\label{eq:evdef}
w_{\cE}(\ob,h) = \frac{\mu_h(\ob)}{\sum_{h'\in\cH}
\mu_{h'}(\ob)}.
\end{equation}
The measure $w_{\cE}$ always lies between 0 and 1; intuitively, if
$w_{\cE}(\ob,h) = 1$, then $\ob$ fully confirms $h$ (i.e., $h$ is
certainly true if $\ob$ is observed), while if $w_{\cE}(\ob,h) = 0$,
then $\ob$ disconfirms $h$ (i.e., $h$ is certainly false if $\ob$ is
observed). Moreover, for each fixed observation $\ob$ for which
$\sum_{h\in\cH}\mu_h(\ob)>0$, $\sum_{h\in\cH} w_{\cE}(\ob, h)=1$, and
thus the weight of evidence $w_\cE$ looks like a probability measure
for each $\ob$. While this has some useful technical consequences, one
should not interpret $w_\cE$ as a probability measure. Roughly
speaking, the weight $w_{\cE}(\ob,h)$ is the likelihood that $h$ is
the right hypothesis in the light of observation $\ob$.%
\footnote{We could have taken the log of the ratio to make
$w_{\cE}$ parallel the log-likelihood ratio $l$ defined
earlier, but there are technical advantages in having the weight of
evidence be a number between 0 and 1.} 
The advantages of $w_{\cE}$ over other known measures of
confirmation are that (a) it is applicable when 
we are not given a prior probability distribution on the hypotheses, 
(b) it is applicable when
there are more than two competing hypotheses, and (c) it has a fairly
intuitive probabilistic interpretation.

An important problem in
\emph{statistical inference} \cite{r:casella01} 
is that of
choosing the best parameter (i.e.,
hypothesis) that explains observed data.
When there is no prior on 
the parameters, the ``best'' parameter is 
typically taken to be the one that
maximizes the likelihood of the data given that parameter. 
Since $w_{\cE}$ is just a normalized likelihood function,
the parameter that maximizes the likelihood will also
maximize $w_{\cE}$.
Thus,
if all we are interested in is maximizing likelihood, there is no
need to normalize the evidence as we do.
We return to the issue of normalization in Section~\ref{s:normalized}.%
\footnote{Another 
representation of evidence that has similar characteristics to $w_{\cE}$
is Shafer's original representation of evidence via belief functions
\cite{r:shafer76},  
defined as
\[w^S_{\cE}(\ob,h) = \frac{\mu_h(\ob)}{\max_{h\in\cH}
\mu_h(\ob)}.\] This measure  
is known in statistical hypothesis testing as the \emph{generalized
likelihood-ratio statistic}. It is another generalization of the
log-likelihood ratio measure $l$.  
The main difference between $w_{\cE}$ and $w^S_{\cE}$ is
how they behave when one considers the combination of evidence,
which we discuss later in this section.
As Walley \citeyear{Walley87} and Halpern and Fagin \citeyear{HF2}
point out, $w_{\cE}$ gives more intuitive results in this case.
We remark that the parameter (hypothesis) that maximized likelihood also
maximizes $w^S_{\cE}$, so $w^S_{\cE}$ can also be used in statistical
inference.}  
Note that if $\cH=\{h_1,h_2\}$, then $w_{\cE}$ in some sense
generalizes the log-likelihood ratio measure. More precisely, for a
fixed observation $\ob$, $w_{\cE}(\ob, \cdot)$ induces the same
relative order on hypotheses as $l(\ob, \cdot)$,
and for a fixed hypothesis $h$, $w_{\cE}(\cdot, h)$ induces the same 
relative order on observations as $l(\cdot, h)$.
\pro\label{p:induces-same}
For all $\ob$, we have $w_{\cE}(\ob,h_i)\geq
w_{\cE}(\ob,h_{3-i})$ if and  only if 
$l(\ob,h_i)\ge l(ob,h_{3-i})$, for $i = 1, 2$, and for all $h$, $\ob$,
and $\ob'$, we have  $w_{\cE}(\ob,h)\geq w_{\cE}(\ob',h)$ if and only
if $l(\ob,h)\geq l(\ob',h)$.  
\epro

Although $w_{\cE}(\ob, \cdot)$ behaves like a probability measure on
hypotheses for every observation $\ob$,
one should not think of it as a probability; 
the weight of evidence of a combined hypothesis, for instance, is not
generally the sum of the weights of the individual hypotheses
\cite{r:halpern05b}. 
Rather, $w_{\cE}(\ob,\cdot)$ is an encoding of evidence.
But what is evidence?
Halpern and Fagin \citeyear{HF2} have suggested that evidence can be
thought of as a \emph{function}
mapping a prior probability on the hypotheses to a posterior
probability, based on the observation made.
There is a precise sense in which $w_{\cE}$ can be viewed as a
function 
that maps a prior probability $\mu_0$ on the
hypotheses $\cH$ to 
a posterior probability $\mu_\ob$ based on observing $\ob$,
by applying Dempster's Rule of Combination \cite{r:shafer76}.  
That is,
\begin{equation}\label{eq:update}
 \mu_\ob = \mu_0 \oplus w_{\cE}(\ob,\cdot),
\end{equation}
where $\oplus$ combines two probability distributions on $\cH$ to get a new
probability distribution on $\cH$ defined as follows:
\[ (\mu_1\oplus\mu_2)(H) = \frac{\sum_{h\in H}\mu_1(h)\mu_2(h)}
                                {\sum_{h\in \cH}\mu_1(h)\mu_2(h)}.\]
(Dempster's Rule of Combination is used to combine belief
functions.  The definition of $\oplus$ is more complicated when
considering arbitrary belief functions, but in the special case where
the belief functions are in fact  probability measures, it takes the
form we give here.) 

Bayes' Rule is the standard way of updating a prior probability
based on an observation, but it is only applicable when we have a
joint probability distribution on both the hypotheses and the
observations
(or, equivalently, a prior on hypotheses together with the likelihood
functions $\mu_h$ for $h \in \cH$), something which we 
do not want to assume we are given. In particular, while we are willing
to assume that we are given the likelihood functions, we are not willing to
assume that we are given a prior on hypotheses.
Dempster's Rule of Combination essentially ``simulates'' the effects of
Bayes' Rule. 
The relationship between Dempster's Rule and Bayes' Rule is made precise
by the following well-known theorem.
\pro\label{p:update}
{\rm\cite{HF2}}
Let $\cE=(\cH,\cO,\bmu)$ be an evidence
space. Suppose that $P$ is a probability on $\cH\times \cO$ such that 
$P(\cH\times\{\ob\} \mid \{h\}\times\cO)=\mu_h(\ob)$ for all $h\in\cH$ and
all $\ob\in\cO$. 
Let $\mu_0$ be the probability on $\cH$ induced by marginalizing
$P$; that is, 
$\mu_0(h)=P(\{h\}\times\cO)$. 
For $\ob \in \cO$, let $\mu_{\ob}=\mu_0\oplus w_{\cE}(\ob,\cdot)$.  Then 
$\mu_{\ob}(h) = P(\{h\}\times\cO \mid \cH\times\{\ob\})$.
\epro
In other words, when we do have a joint probability on the hypotheses
and observations, then Dempster's Rule of Combination gives us the
same result as a straightforward application of Bayes' Rule.

\xam\label{x:coins}
To get a feel for how this measure of evidence can be used, consider a
variation of the two-coins example in the introduction. Assume that
the coin chosen by Alice is either double-headed or fair,
and consider sequences of a hundred tosses of that coin. 
Let $\cO = \{m: 0 \le m \le 100\}$ (the number of heads observed), and
let $\cH = \{F,D\}$, where
$F$ is ``the coin is fair'', and $D$ is ``the coin is
double-headed''. The probability spaces associated 
with the hypotheses
are generated by the following probabilities for simple
observations $m$:
\[ \mu_F(m) = \frac{1}{2^{100}}{100 \choose m} \qquad  \mu_D(m) = \left\{\begin{array}{ll}
   1 & \mbox{if $m=100$}\\
   0 & \mbox{otherwise.}
                       \end{array}\right.\]
(We extend by additivity to the whole set $\cO$.) 
Take $\cE=(\cH,\cO,\bmu)$, where $\bmu(F)=\mu_F$ and $\bmu(D)=\mu_D$.
For any observation $m\neq 100$, the weight in favor of $F$ is given
by 
\[ w_{\cE}(m,F) = \frac{\frac{1}{2^{100}}{100 \choose
m}}{0+\frac{1}{2^{100}}{100 \choose m}} = 1,\]
which 
means that the support of $m$ is unconditionally
provided to $F$; indeed, any such sequence of tosses cannot appear with
the double-headed coin. 
Thus, if $m \ne 100$, we get that
\[ w_{\cE}(m,D) = \frac{0}{0+\frac{1}{2^{100}}{100 \choose m}} = 0.\]
What happens when the hundred coin tosses are all heads? 
It is straightforward to check that 
\[ w_{\cE}(100,F) = \frac{\frac{1}{2^{100}}}{1+\frac{1}{2^{100}}} =
\frac{1}{1+2^{100}}   \qquad 
 w_{\cE}(100,D) = \frac{1}{1+\frac{1}{2^{100}}} = \frac{2^{100}}{1+2^{100}};  \]
this time there is overwhelmingly more evidence in favor of $D$ than $F$.

Note that we have not assumed any prior probability.  
Thus, we cannot talk about the probability that the coin
is fair or double-headed. What we have is a quantitative assessment of the
evidence in favor of one of the hypotheses.  
However, if we assume a prior probability $\alpha$ on the coin being
fair and $m$ heads are observed after 100 tosses, then the probability
that the coin is fair is 1 if $m \ne 100$; if $m = 100$ then, 
applying the rule
of combination, the posterior probability of the coin being fair is 
$\alpha /(\alpha  + (1-\alpha)2^{100})$.
\exam

Can we characterize weight functions using a small number of
properties? More precisely, given sets $\cH$ and $\cO$, and a function
$f$ from $\cO\times\cH$ to $[0,1]$, are there properties of $f$ that 
ensure that there are likelihood functions $\bmu$ such
that $f=w_\cE$
for $\cE=(\cH,\cO,\bmu)$? 
As we saw earlier, for a fixed observation $\ob$, $f$
essentially 
acts like a probability measure on $\cH$. However, this is not
sufficient to guarantee that $f$ is a weight function. Consider the
following example, with $\cO=\{\ob_1,\ob_2\}$ and
$\cH=\{h_1,h_2,h_3\}$:
\[\begin{array}{rclcrcl}
f(\ob_1,h_1) & = & 1/4 & \quad & f(\ob_2,h_1) & = & 1/4\\
f(\ob_1,h_2) & = & 1/4 & \quad & f(\ob_2,h_2) & = & 1/2\\
f(\ob_1,h_3) & = & 1/2 & \quad & f(\ob_2,h_3) & = & 1/4.
  \end{array}\]
It is straightforward to check that $f(\ob_1,\cdot)$ and $f(\ob_2,\cdot)$ are
probability measures on $\cH$, but that there is no evidence space
$\cE=(\cH,\cO,\bmu)$ such that
$f=w_{\cE}$. 
Indeed, assume that we do have such
$\mu_{h_1},\mu_{h_2},\mu_{h_3}$. By the definition of weight of
evidence, and the fact that $f$ is that weight of evidence, we get the 
following system of equations:
\[ \begin{array}{l}
\frac{\mu_{h_1}(\ob_1)}
     {\mu_{h_1}(\ob_1)+\mu_{h_2}(\ob_1)+\mu_{h_3}(\ob_1)} = 1/4\\
\frac{\mu_{h_2}(\ob_1)}
     {\mu_{h_1}(\ob_1)+\mu_{h_2}(\ob_1)+\mu_{h_3}(\ob_1)} = 1/4\\
\frac{\mu_{h_3}(\ob_1)}
     {\mu_{h_1}(\ob_1)+\mu_{h_2}(\ob_1)+\mu_{h_3}(\ob_1)} = 1/2
   \end{array}\quad
   \begin{array}{l}
\frac{\mu_{h_1}(\ob_2)}
     {\mu_{h_1}(\ob_2)+\mu_{h_2}(\ob_2)+\mu_{h_3}(\ob_2)} = 1/4\\
\frac{\mu_{h_2}(\ob_2)}
     {\mu_{h_1}(\ob_2)+\mu_{h_2}(\ob_2)+\mu_{h_3}(\ob_2)} = 1/2\\
\frac{\mu_{h_3}(\ob_2)}
     {\mu_{h_1}(\ob_2)+\mu_{h_2}(\ob_2)+\mu_{h_3}(\ob_2)} = 1/4.
   \end{array}\]
It is now immediate that there exist $\alpha_1$ and $\alpha_2$ such that
$\mu_{h_i}(\ob_j) = \alpha_j  f(\ob_j, h_i)$, for $i = 1, 2, 3$.  Indeed,
$\alpha_j =\mu_{h_1}(\ob_j)+\mu_{h_2}(\ob_j)+\mu_{h_3}(\ob_j)$, for
$j=1, 2$.  
Moreover, since $\mu_{h_i}$ is a probability measure, we must have that 
\[\mu_{h_i}(\ob_1) + \mu_{h_i}(\ob_2) = \alpha_1 f(\ob_1,h_i) + \alpha_2
f(\ob_2, h_i) = 1,\]
for $i = 1, 2, 3$.
Thus, $$\alpha_1/4 + \alpha_2/4  = \alpha_1/4 + \alpha_2/2 = \alpha_1/2
+ \alpha_4/4 = 1.$$
These constraints are easily seen to be unsatisfiable.

This argument generalizes to arbitrary functions
$f$; thus, a necessary condition for $f$ to be a weight function is
that there exists $\alpha_i$ for each observation $\ob_i$ such that
$\mu_h(\ob_i)=\alpha_i f(\ob_i,h)$ for each hypothesis $h$ is a
probability measure, that is, $\alpha_1 f(\ob_1,h)+\dots+\alpha_k
f(\ob_k,h)=1$. In fact, when combined with the constraint that
$f(\ob,\cdot)$ is a probability measure for a fixed $\ob$, this
condition turns out to be sufficient, as the following theorem
establishes.
\thm\label{t:characterization}
Let $\cH=\{h_1,\dots,h_m\}$ and $\cO=\{\ob_1,\dots,\ob_n\}$, and let
$f$ be a real-valued function with domain $\cO\times\cH$ such 
that $f(\ob,h)\in[0,1]$. Then there exists an evidence space
$\cE=(\cH,\cO,\bmu)$ such that $f=w_{\cE}$ if
and only if $f$ satisfies the following properties:
\begin{axiomlist}
\item[\rm\axiom{WF1}.] For every $\ob\in\cO$, $f(\ob,\cdot)$ is a
probability measure on $\cH$.
\item[\rm\axiom{WF2}.] There exists $x_1,\dots,x_n > 0$ such that, for all
$h\in\cH$, $\sum_{i=1}^{n}f(\ob_i,h)x_i=1$.
\end{axiomlist}
\ethm

This characterization is fundamental to the completeness of the
axiomatization of the logic we introduce in the next section. 
The characterization is complicated by the fact that the weight of
evidence is essentially a normalized likelihood:  the likelihood
of an observation given a particular hypothesis is normalized using
the sum of all the likelihoods of that observation, for all possible
hypotheses. One consequence of this, as we already mentioned above, is
that the weight of evidence is always between 0 and 1, and
superficially behaves like a probability measure. 
In Section~\ref{s:normalized}, we examine the issue of normalization
more carefully, and describe the changes to our framework that would
occur were we to take unnormalized likelihoods as weight of evidence.
Let $\cE=(\cH,\cO,\bmu)$ be an evidence space. 
Let $\cO^*$ be the set of sequences of observations
$\<\ob^1,\dots,\ob^k\>$ over $\cO$.\footnote{We use superscript rather
than subscripts to index observations in a sequence so that these
observations will not be confused with the basic observations
$\ob_1,\dots,\ob_n$ in $\cO$.}
Assume that the observations are independent, that
is, for each basic hypothesis $h$, 
take
$\mu^*_h(\<\ob^1,\dots,\ob^k\>)$, the probability of observing a
particular sequence of observations given $h$, 
to be
$\mu_h(\ob^1)\cdots\mu_h(\ob^k)$, the product of the probability of
making each observation in the sequence.  Let
$\cE^*=(\cH,\cO^*,\bmu^*)$.
With this assumption, it is well known 
that Dempster's Rule of Combination can be
used to combine evidence in this setting; that is,
\[ w_{\cE^*}(\<\ob^1,\dots,\ob^k\>,\cdot) = w_{\cE}(\ob^1,\cdot)\oplus\dots\oplus
w_{\cE}(\ob^k,\cdot)\] 
\cite[Theorem~4.3]{HF2}. 
It is an easy exercise to check that the weight provided by the
sequence of observations $\<\ob^1,\dots,\ob^k\>$ can be expressed in
terms of the weight of the individual observations:
\begin{equation}\label{e:decomposition}
w_{\cE^*}(\<\ob^1,\dots,\ob^k\>,h) = \frac{w_{\cE^*}(\ob^1,h)\cdots
w_{\cE^*}(\ob^k,h)}{\sum_{h'\in\cH}w_{\cE^*}(\ob^1,h')\cdots w_{\cE^*}(\ob^k,h')}.
\end{equation}
If we let $\mu_0$ be a prior probability on the hypotheses, and
$\mu_{\<\ob^1,\dots,\ob^k\>}$ be the probability on the hypotheses
after observing $\ob^1,\dots,\ob^k$, we can verify that
\[\mu_{\<ob^1,\dots,\ob^k\>} = \mu_0\oplus
w_{\cE^*}(\<\ob^1,\dots,\ob^k\>,\cdot).\] 

\xam\label{x:coinsseq}
Consider a variant of Example~\ref{x:coins}, where we take the coin 
tosses as individual observations, rather than the number of heads
that turn up in one hundred coin tosses. As before, assume that the
coin chosen by Alice is either double-headed or fair. Let
$\cO=\{H,T\}$, the result of an individual coin toss, where $H$ is
``the coin landed heads'' and $T$ is ``the coin landed tails''. Let
$\cH=\{F,D\}$, where $F$ is ``the coin is fair'', and $D$ is ``the
coin is double-headed''. 
Let $\cE^*=(\cH,\cO^*,\bmu^*)$.
The probability measure $\mu^*_h$ associated with the 
hypothesis $h$ are generated by the following probabilities for simple
observations:
\[ \mu_F(H) = \frac{1}{2} \qquad  \mu_D(H) = 1.\]
Thus, for example, $\mu^*_F(\<H, H, T, H\>) = 1/16$, 
$\mu^*_D(\<H, H, H\>) = 1$, and $\mu^*_H(\<H, H, T, H\>) = 0$.

We can now easily verify
results similar to those that were obtained in
Example~\ref{x:coins}. For instance, the weight of observing $T$ in
favor of $F$ is given by 
\[ w_{\cE^*}(T,F) = \frac{\frac{1}{2}}{0+\frac{1}{2}} = 1,\]
which again indicates that observing $T$ provides unconditional support
to $F$; a double-headed coin cannot land tails. 

How about sequences of observations? 
The weight provided by the sequence $\<\ob^1,\dots,\ob^k\>$ for
hypothesis $h$ is
given by Equation (\ref{e:decomposition}).
Thus, if $\sequ{H}=\<H,\dots,H\>$, a sequence of a hundred coin
tosses, we can check that
\[ w_{\cE^*}(\sequ{H},F) = \frac{\frac{1}{2^{100}}}{1+\frac{1}{2^{100}}} =
\frac{1}{1+2^{100}}
  \qquad
   w_{\cE^*}(\sequ{H},D) = \frac{1}{1+\frac{1}{2^{100}}} = \frac{2^{100}}{1+2^{100}}.\]
Unsurprisingly, this is the same result as in Example~\ref{x:coins}.
\exam

\section{Reasoning about Evidence}\label{s:logic}

We introduce a logic $\Lfoev$ for reasoning about evidence, inspired by
a logic introduced in FHM
for reasoning about probability.
The logic lets us reason about the weight of evidence of observations
for hypotheses; moreover, to be able to talk about the relationship
between prior probabilities, evidence, and posterior probabilities, we
provide operators to reason about the prior and posterior
probabilities of hypotheses. 
We remark that up to now we have been somewhat agnostic about whether
the priors exist but are not given (or not known) or whether the prior
does not exist at all.  It is beyond the scope of this paper to enter
the debate about whether it always appropriate to assume the existence
of a prior.   
Although the definition of evidence makes sense even if
the priors does not exist, our logic implicitly assumes that there are
priors (although they may not be known), since we provide operators for
reasoning about the prior.  We make use of these operators in some of
the examples below.  However, the fragment of the logic that does
not use these operators is appropriate for prior-free reasoning.

The logic has both propositional features
and first-order features.  
We take  the probability of propositions and the
weight of evidence of observations for hypotheses, and view
probability and evidence as propositions, 
but
we allow first-order quantification over numerical
quantities, such as probabilities and evidence.  The logic essentially
considers two time periods, which can be thought of as the time before
an observation is made and the time after an observation is made.  
In this section, we assume that exactly one observation is made. (We
consider sequences of observations in Section~\ref{s:dynamic}.) Thus,
we can talk of the probability of a formula $\phi$ before an
observation is made, denoted $\Pr^0(\phi)$, the probability of $\phi$
after the observation, denoted $\Pr(\phi)$, and the evidence
provided by 
the observation $\ob$ for an
hypothesis $h$, denoted $\We(\ob,h)$.  Of course, we want to be able to
use the logic to relate all these quantities.

Formally, we start with two 
finite
sets of primitive propositions,
$\Phih=\{h_1,\dots,h_{\nh}\}$ representing the hypotheses, and
$\Phio=\{\ob_1,\dots,\ob_{\no}\}$ representing the observations.
Let 
$\Lh(\Phih)$ 
be the propositional sublanguage of
\emph{hypothesis formulas} obtained by taking primitive propositions
in $\Phih$ and closing off under negation and conjunction; we use
$\rho$ to range over formulas of that sublanguage. 

A {\em basic term\/} has the form $\Pr^0(\rho)$,
$\Pr(\rho)$, or $\We(\ob,h)$, where $\rho$ is an hypothesis
formula, $\ob$ is an observation, and $h$ is an hypothesis.  
As we said,
we interpret $\Pr^0(\rho)$ as the prior probability of $\rho$, 
$\Pr(\rho)$ as the posterior probability of $\rho$, and $\We(\ob,h)$
as the weight of evidence of observation $\ob$ for hypothesis $h$.
It may seem strange that we allow the language to talk about the prior
probability of hypotheses, although we have said that we do not want to
assume that the prior is known.  We could, of course, simplify the
syntax so that it did not include formulas of the form $\Pr^0(\rho)$ or
$\Pr(\rho)$.  The advantage of having them is that, even if the prior is
not known, given our view of evidence as a function from priors to
posteriors, we can make statements such as ``if the prior probability of
$h$ is $2/3$, $\ob$ is observed, and the weight of evidence of $\ob$ for
$h$ is $3/4$, then the posterior probability of $h$ is $6/7$; this is
just $$\Pr^0(h) = 1/2 \land \ob \land \We(\ob,h) = 3/4 \rimp \Pr(h) = 6/7.$$

A {\em polynomial term\/} has the  
form $t_1 + \dots + t_n$, where each term $t_i$ is a product of
integers, basic terms, and variables (which range over the
reals). A {\em polynomial inequality formula\/} has the form $p \ge
c$, where $p$ is a polynomial term and $c$ is an integer.
Let  
$\Lfoev(\Phih,\Phio)$ be the 
language obtained by starting out with the primitive propositions in
$\Phih$ and $\Phio$ and polynomial inequality formulas, and closing off under
conjunction, negation, and first-order quantification.
Let $\truep$ be an abbreviation for an arbitrary propositional
tautology involving only hypotheses, such as $h_1\lor\neg h_1$;  let 
$\falsep$ be an abbreviation for $\neg\truep$. With this definition,
$\truep$ and $\falsep$ can be considered as part of the sublanguage
$\Lh(\Phih)$. 

It should be clear that while we allow only integer coefficients to
appear in polynomial terms, we can in fact express polynomial terms
with rational coefficients by crossmultiplying. 
For instance, $\frac{1}{3}\Pr(\rho)+\frac{1}{2}\Pr(\rho')\ge 1$ can be 
represented by the polynomial inequality formula
$2\Pr(\rho)+3\Pr(\rho')\ge 6$.
While there is no 
difficulty in giving a semantics to polynomial terms that use
arbitrary real coefficients, we need the restriction to integers in
order to make use of results from the theory of real closed fields in
both the axiomatization of Section~\ref{s:axiomatization} and the
complexity results of Section~\ref{s:decision}. 

We use obvious abbreviations where needed, such as $\phi\lor\psi$ for
$\neg(\neg\phi\land\neg\psi)$, $\phi\rimp\psi$ for $\neg\phi\lor\psi$, 
$\exists x\phi$ for $\neg\forall x(\neg \phi)$, 
$\Pr(\phi)-\Pr(\psi)\geq c$ for $\Pr(\phi)+(-1)\Pr(\psi)\geq
c$, $\Pr(\phi)\geq \Pr(\psi)$ for $\Pr(\phi)-\Pr(\psi)\geq 0$,
$\Pr(\phi)\leq c$ for $-\Pr(\phi)\geq -c$, $\Pr(\phi)<
c$ for $\neg(\Pr(\phi)\geq c)$, and $\Pr(\phi)=c$ for
$(\Pr(\phi)\geq c)\land(\Pr(\phi)\leq c)$ (and analogous
abbreviations for inequalities involving $\Pr^0$ and $\We$).

\xam\label{x:coins2}
Consider again the situation given in Example~\ref{x:coins}.
Let $\Phio$, the observations, consist of 
primitive propositions of the form $\mathsf{heads}[m]$, where $m$ is
an integer with $0\leq m\leq 100$, indicating
that $m$ heads out of 100 tosses have appeared. Let $\Phih$
consist of the two primitive propositions $\mathsf{fair}$ and
$\mathsf{doubleheaded}$. The computations in Example~\ref{x:coins}
can be written as follows:
\[ \We(\mathsf{heads}[100],\mathsf{fair})=1/(1+2^{100})\land
   \We(\mathsf{heads}[100],\mathsf{doubleheaded})=2^{100}/(1+2^{100}).\]
We can also capture the fact that the weight of evidence of an
observation maps a prior probability into a posterior 
probability by Dempster's Rule of Combination.
For example, the
following formula captures the update of the prior probability
$\alpha$ of the hypothesis $\mathsf{fair}$ upon observation of a
hundred coin tosses landing heads:
\[
\Pr^0(\mathsf{fair})=\alpha\land\We(\mathsf{heads}[100],\mathsf{fair})=1/(1+2^{100})\rimp 
\Pr(\mathsf{fair})=\alpha/(\alpha+(1-\alpha)2^{100}).\]
We 
develop a deductive system to derive such conclusions in the next
section.  
\exam

Now we consider the semantics.  A formula is interpreted in a world
that specifies which hypothesis is true and which observation was
made, as well as an evidence space to interpret the weight of evidence
of observations and a probability distribution on the hypotheses to
interpret prior probabilities and talk about updating based on
evidence. (We do not need to include a posterior probability
distribution, since it can be computed from the prior and the weights of
evidence using Equation \eqref{eq:update}.) An \emph{evidential world}
is a tuple $w=(h,\ob,\mu,\cE)$, where $h$ is a hypothesis, $\ob$ is an
observation, $\mu$ is a probability distribution on $\Phih$, and
$\cE$ is an evidence space over $\Phih$ and $\Phio$.

To interpret propositional formulas in $\Lh(\Phih)$, 
we associate with each hypothesis formula $\rho$ a set
$\intension{\rho}$ of hypotheses, by induction on the structure of
$\rho$:
\begin{eqnarray*}
\intension{h} & = & \{h\}\\
\intension{\neg\rho} & = & \Phih-\intension{\rho}\\
\intension{\rho_1\land\rho_2} & = &
\intension{\rho_1}\inter\intension{\rho_2}.
\end{eqnarray*}
To interpret 
first-order formulas that may contain variables, we need a valuation
$v$ that assigns a real number to every variable. 
Given an evidential world $w = (h,\ob,\mu,\cE)$ and a valuation $v$,
we 
assign to a polynomial term $p$ a real number $\val{p}{w,v}$
in a straightforward way:  
\begin{align*}
\val{x}{w,v} & =  v(x)\\
\val{a}{w,v} & =  a\\
\val{\Pr^0(\rho)}{w,v} & =  \mu(\intension{\rho})\\ 
\val{\Pr(\rho)}{w,v} & = 
    (\mu\oplus w_{\cE}(\ob,\cdot))(\intension{\rho})\\ 
\val{\We(\ob',h')}{w,v} & = 
w_{\cE}(\ob',h')\\
\val{t_1 t_2}{w,v} & =  \val{t_1 }{w,v} \times
\val{t_2}{w,v}\\
\val{p_1 + p_2}{w,v} & =  \val{p_1}{w,v} + \val{p_2}{w,v}.
\end{align*}
Note that, to interpret $\Pr(\rho)$, the posterior probability of
$\rho$ after having observed $\ob$ (the observation at world $w$), we
use Equation \eqref{eq:update}, which says that the posterior is
obtained by combining the prior probability $\mu$ with
$w_{\cE}(\ob,\cdot)$. 

We define what it means for a formula $\phi$ to be true (or satisfied)
at an evidential world $w$ under valuation $v$, written
$(w,v)\sat\phi$, as follows:  
\begin{itemize}
\item[]$(w,v)\sat h$ if $w=(h,\ob,\mu,\cE)$ for some $\ob$, $\mu$, $\cE$ 
\item[]$(w,v)\sat\ob$ if $w=(h,\ob,\mu,\cE)$ for some $h$, $\mu$, $\cE$
\item[]$(w,v)\sat\neg\phi$ if $(w,v)\not\sat\phi$
\item[]$(w,v)\sat\phi\land\psi$ if $(w,v)\sat\phi$ and
$(w,v)\sat\psi$
\item[]$(w,v)\sat p \ge c$ if $\val{p}{w,v} \ge c$
\item[]$(w,v)\sat\forall x\phi$ if $(w,v')\sat\phi$ for
all $v'$ that agree with $v$ on all variables but $x$. 
\end{itemize}

If $(w,v)\sat\phi$ is true for all $v$, we write simply
$w\sat\phi$. It is easy to check that if $\phi$ is a closed
formula (that is, one with no free variables),
then $(w,v)\sat\phi$ if and only if
$(w,v')\sat\phi$, for all $v,v'$. Therefore, given a closed
formula $\phi$, if $(M,w,v)\sat\phi$, then in fact
$w\sat\phi$.  We will typically be concerned only with closed
formulas.
Finally, if $w\sat\phi$ for all evidential worlds $w$, we write
$\sat\phi$ and say
that
$\phi$ is valid. In the next section, we will characterize
axiomatically all the valid formulas of the logic.

\xam\label{x:valid}
The following formula is valid, that is, true in all evidential
worlds: 
\[ \sat (\We(\ob,h_1)=2/3 \land \We(\ob,h_2)=1/3) \rimp (\Pr^0(h_1)\ge
1/100 \land \ob)\rimp \Pr(h_1)\ge 2/101.\]
In other words, at all evidential worlds where the weight of evidence
of observation $\ob$ for hypothesis $h_1$ is 2/3 and the weight of
evidence of observation $\ob$ for hypothesis $h_2$ is 1/3, it must be
the case that if the prior probability of $h_1$ is at least 1/100 and
$\ob$ is actually observed, then the posterior probability of $h_1$ is
at least 2/101. This shows the extent to which we can reason about the
evidence independently of the prior probabilities. 
\exam

The logic imposes no restriction on the prior probabilities to be
used in the models. This implies, for instance, that the 
formula 
\[  \mathsf{fair} \rimp \Pr^0(\mathsf{fair})=0\]
is satisfiable: there exists an evidential world $w$ such that the
formula is true at $w$. In other words, it is consistent for an
hypothesis to be true, despite the prior probability of it being true
being 0. It is a simple matter to impose a restriction on the models
that they be such that if $h$ is true at a world, then $\mu(h)>0$ for
the prior $\mu$ at that world.

We conclude this section with some remarks concerning 
the semantic model.   Our semantic model implicitly assumes that
the prior 
probability is known and that the likelihood functions (i.e., the
measures $\mu_h$) are known.  Of course, in many situations there 
will be uncertainty about both.  Indeed, our motivation for focusing on
evidence is precisely to deal with situations where the prior is not
known.  Handling uncertainty about the prior is  
easy in our framework, since our notion of evidence is independent of
the prior on hypotheses.  It is straightforward to extend our model by
allowing a set of possible worlds, with a different prior in each, but
using the same evidence space for all of them.  We
can then extend the logic with a knowledge operator, where a statement
is known to be true if it is true in all the worlds.  This allows us to
make statements like ``I know that the prior on hypothesis $h$ is between
$\alpha$ and $\beta$.  Since observation $\ob$ provides evidence $3/4$ for 
$h$, I know that the posterior on $h$ given $\ob$ is 
between $(3\alpha)/(2\alpha + 1)$ and $(3 \beta)/(2\beta + 1)$.''

Dealing with uncertainty about the likelihood functions is somewhat more
subtle.  To understand the issue, suppose that one of two
coins will be chosen and tossed.  The bias of coin 1 (i.e., the
probability that coin 1 lands heads) is between $2/3$ and
$3/4$; the bias of coin 2 is between $1/4$ and $1/3$.  Here there 
is uncertainty about the probability that coin 1 will be picked (this is
uncertainty about the prior) and there is uncertainty about the bias of
each coin (this is uncertainty about the likelihood functions).
The problem here is that, to deal with this, we must consider possible
worlds where there is a possibly different evidence space in each world.
It is then not obvious how to define weight of evidence.  We explore
this issue in more detail in a companion paper \cite{r:halpern05b}.

\section{Axiomatizing Evidence}\label{s:axiomatization}

In this section we present a sound and complete axiomatization 
$\AX$
for our logic. 
The axiomatization can be divided into four parts.
The first part, consisting of the following axiom and inference rule,
accounts for  
first-order
reasoning:
\begin{axiomlist}
\item[\axiom{Taut}.] All 
substitution
instances of valid formulas of first-order
logic with equality.
\item[\axiom{MP}.] From $\phi$ and $\phi\rimp\psi$ infer $\psi$.
\end{axiomlist}
Instances of \axiom{Taut} include, for example, all formulas of the
form $\phi\lor\neg\phi$, where $\phi$ is an arbitrary formula of the
logic. 
It also includes formulas such as $(\forall x\phi)\riff\phi$ if $x$ is
not free in $\phi$. In particular, $(\forall x(h))\riff h$ for
hypotheses in $\Phih$, and similarly for observations in $\Phio$. 
Note that \axiom{Taut} includes all \emph{substitution} instances of
valid formulas of first-order logic with equality; in other words, any
valid formula of first-order logic with equality where free variables
are replaced with arbitrary terms of our language (including
$\Pr^0(\rho)$, $\Pr(\rho)$, $\We(\ob,h)$) is an instance of
\axiom{Taut}. 
Axiom \axiom{Taut} can be replaced by a sound and complete
axiomatization for first-order logic with equality, as given, for
instance, in Shoenfield \citeyear{Shoen} or Enderton
\citeyear{r:enderton72}. 

The second set of axioms accounts for reasoning about polynomial
inequalities, by relying on the theory of real closed fields:
\begin{axiomlist}
\item[\axiom{RCF}.] All instances of formulas valid in real closed
fields (and, thus, true about the reals), 
with nonlogical symbols $+$, $\cdot$, $<$, $0$, $1$, $-1$, $2$,
$-2$, $3$, $-3$, \dots.
\end{axiomlist}
Formulas that are valid in real closed fields
include, for example, the fact
that addition on the reals is associative, $\forall x\forall y\forall
z((x+y)+z = x+(y+z))$, that $1$ is the identity for multiplication,
$\forall x(x\cdot1 = x)$,  
and formulas relating the constant symbols, such as $k = 1 + \dots + 1$
($k$ times) and $-1 + 1 = 0$.
As for \axiom{Taut}, we could replace \axiom{RCF} by a sound and
complete axiomatization for real closed fields
\cite<cf.>{r:fagin90,Shoen,Tar1}.

The third set of axioms essentially captures the fact that there is a
single hypothesis and a single observation that holds per state. 
\begin{axiomlist}
\item[\axiom{H1}.] $h_1\lor\dots\lor h_{\nh}$.
\item[\axiom{H2}.] $h_i\rimp\neg h_j$ if $i\ne j$. 
\item[\axiom{O1}.] $\ob_1\lor\dots\lor \ob_{\no}$.
\item[\axiom{O2}.] $\ob_i\rimp\neg \ob_j$ if $i\ne j$. 
\end{axiomlist}
These axioms illustrate a subtlety of our logic. Like most
propositional logics, ours is parameterized by primitive
propositions, in our case, $\Phih$ and $\Phio$. However, while
axiomatizations for propositional logics typically do not depend on
the exact set of primitive propositions, ours does. Clearly, axiom
\axiom{H1} is sound only if the hypothesis primitives are exactly
$h_1,\dots,h_{\nh}$. Similarly, axiom \axiom{O1} is sound only if the
observation primitives are exactly $\ob_1,\dots,\ob_{\no}$. It is
therefore important for us to identify the primitive propositions when 
talking about the axiomatization $\AX$. 

The last set of axioms concerns reasoning about probabilities and
evidence proper. The axioms for probability are taken from FHM. 
\begin{axiomlist}
\item[\axiom{Pr1}.] $\Pr^0(\truep)=1$.
\item[\axiom{Pr2}.] $\Pr^0(\rho)\geq 0$.
\item[\axiom{Pr3}.] $\Pr^0(\rho_1\land\rho_2)+\Pr^0(\rho_1\land\neg\rho_2)=\Pr^0(\rho_1)$.
\item[\axiom{Pr4}.] $\Pr^0(\rho_1)=\Pr^0(\rho_2)$ if
$\rho_1\riff\rho_2$ is a propositional tautology.
\end{axiomlist}
Axiom \axiom{Pr1} simply says that the event $\truep$ has probability
$1$. Axiom \axiom{Pr2} says that probability is nonnegative. Axiom
\axiom{Pr3} captures finite additivity. It is not possible to express
countable additivity in our logic. On the other hand, 
just as in FHM,
we do not need
an axiom for
countable additivity. Roughly speaking, as we establish in the next
section, if a formula is satisfiable at all, it is satisfiable in a
finite structure. Similar axioms capture posterior probability
formulas:
\begin{axiomlist}
\item[\axiom{Po1}.] $\Pr(\truep)=1$.
\item[\axiom{Po2}.] $\Pr(\rho)\geq 0$.
\item[\axiom{Po3}.] $\Pr(\rho_1\land\rho_2)+\Pr(\rho_1\land\neg\rho_2)=\Pr(\rho_1)$.
\item[\axiom{Po4}.] $\Pr(\rho_1)=\Pr(\rho_2)$ if
$\rho_1\riff\rho_2$ is a propositional tautology.
\end{axiomlist}

Finally, we need axioms to account for the behavior of the evidence
operator $\We$. 
What are these properties?  For one thing, 
the weight function acts 
essentially 
like a probability on
hypotheses, for each fixed observation,
except that we are restricted to taking the weight of evidence of
basic hypotheses only. 
This gives the following 
axioms:
\begin{axiomlist}
\item[\axiom{E1}.] $\We(\ob,h)\geq 0$.
\item[\axiom{E2}.] $\We(\ob,h_1)+\dots+\We(\ob,h_{\nh})=1$. 
\end{axiomlist}

Second, evidence connects the prior and posterior beliefs via Dempster's
Rule of Combination, as in (\ref{eq:update}).  This is captured by the
following axiom.  (Note that, since we do not have division in the
language, we crossmultiply to clear the denominator.)
\begin{axiomlist}
\item[\axiom{E3}.] $\ob\rimp(
\Pr^0(h)\We(\ob,h) = 
\Pr(h)\Pr^0(h_1)\We(\ob,h_1) + \dots +
\Pr(h)\Pr^0(h_{\nh})\We(\ob,h_{\nh})).$
\end{axiomlist}

This is not quite enough.
As we saw in Section~\ref{s:evidence}, 
property \axiom{WF2} in Theorem~\ref{t:characterization} is 
required for a function to be an evidence function. The
following axiom captures \axiom{WF2} in our logic:
\begin{axiomlist}
\item[\axiom{E4}.] $\exists x_1\dots\exists x_{\no}(
\begin{prog}
x_1> 0\land\dots\land x_{\no}> 0 \land 
 \We(\ob_1,h_1)x_1+\dots+\We(\ob_{\no},h_1)x_{\no}=1 \land\\
 \dots \land
 \We(\ob_1,h_{\nh})x_1+\dots+\We(\ob_{\no},h_{\nh})x_{\no}=1). 
\end{prog}$   
\end{axiomlist}
Note that axiom \axiom{E4} is the only axiom that requires
quantification. 
Moreover, axioms \axiom{E3} and \axiom{E4} both depend on $\Phih$ and
$\Phio$. 

As an example, we show that if $h$ and $h'$ are distinct
hypotheses in $\Phih$, then the formula 
\[\neg (\We(\ob,h)=2/3 \land \We(\ob,h')=2/3)\]
is provable. First, by \axiom{RCF}, the following
valid formula of the theory of real closed fields is provable:
\[\forall x\forall y(x=2/3\land y=2/3\rimp x+y>1).\]
Moreover, if $\phi(x,y)$ is any first-order logic formula with two free
variables $x$ and $y$, then 
\[ (\forall x\forall y(\phi(x,y)))\rimp \phi(\We(\ob,h),\We(\ob,h'))
\]
is a substitution instance of a valid formula of first-order logic
with equality, and hence is an instance of \axiom{Taut}. Thus, by
\axiom{MP}, we can prove that 
\[ \We(\ob,h)=2/3 \land \We(\ob,h')=2/3 \rimp
\We(\ob,h)+\We(\ob,h')>1,\]
which is provably equivalent (by \axiom{Taut} and \axiom{MP}) to its
contrapositive 
\[ \We(\ob,h)+\We(\ob,h')\le 1\rimp
\neg(\We(\ob,h)=2/3\land\We(\ob,h')=2/3).\]
By an argument similar to that above, using \axiom{RCF}, \axiom{Taut},
\axiom{MP}, \axiom{E1}, and \axiom{E2}, we can derive 
\[ \We(\ob,h)+\We(\ob,h')\le 1,\]
and by \axiom{MP}, we obtain the desired conclusion:
$\neg(\We(\ob,h)=2/3\land\We(\ob,h')=2/3)$.

\thm\label{t:sound-complete}
$\AX$ is a sound and complete axiomatization for $\Lfoev(\Phih,\Phio)$
with respect to evidential worlds.
\ethm

As usual, soundness is straightforward, and to prove completeness, it
suffices to show that if a formula $\phi$ is consistent with $\AX$, it
is satisfiable in an evidential structure. However, the usual approach
for proving completeness in modal logic, which involves considering
maximal consistent sets and canonical structures does not work. The
problem is that there are maximal consistent sets of formulas that are
not satisfiable. For example, there is a maximal consistent set of
formulas that includes $\Pr(\rho)>0$ and $\Pr(\rho)\leq 1/n$ for $n =
1,2,\dots$. This is clearly unsatisfiable. Our proof follows the
techniques developed in FHM.

To express axiom \axiom{E4}, we needed to have quantification in the
logic.
This is where the fact that our representation of evidence is
normalized has a nontrivial effect on the logic: \axiom{E4} corresponds
to property \axiom{WF2}, which essentially says that a function is a
weight of evidence function if one can find such a normalization
factor.  
An interesting question is whether it is possible to
find a sound and complete axiomatization for the propositional
fragment of our logic (without quantification or variables).
To do this, we need to give
quantifier-free axioms to replace axiom \axiom{E4}.
This amounts to asking whether there is a simpler property than
\axiom{WF2} in Theorem~\ref{t:characterization} that characterizes
weight of evidence functions. This remains an open question.

\section{Decision Procedures}\label{s:decision}

In this section, we consider the decision problem for our logic, that
is, the problem of deciding whether a given formula $\phi$ is
satisfiable. In order to state the problem precisely, however, we need
to deal carefully with the fact that the logic is parameterized by the
 sets $\Phih$ and $\Phio$ of primitive propositions representing
hypotheses and observations. 
In most logics, 
the choice of underlying primitive propositions is essentially
irrelevant.  For example, if a propositional formula $\phi$ that
contains only primitive propositions in some set $\Phi$ is true with
respect to all truth assignments to $\Phi$, then it remains true with
respect to all truth assignments to any set $\Phi' \supseteq \Phi$.
This monotonicity property does not hold here.
For example, 
as we have already observed,
axiom \axiom{H1} clearly depends on the set
of hypotheses and observations; it is no longer valid if the set is
changed. 
The same is true for \axiom{O1}, \axiom{E3}, and \axiom{E4}.

This means that we have to be careful, when stating decision problems,
about the role of $\Phih$ and $\Phio$ in the algorithm. A
straightforward way to deal with this is to assume that the
satisfiability algorithm gets as input $\Phih$, $\Phio$, and a formula
$\phi \in \Lfoev(\Phih,\Phio)$.
Because $\Lfoev(\Phih,\Phio)$ contains the full theory of real
closed fields,
it
is unsurprisingly difficult to decide. 
For our decision procedure, we can use
the exponential-space algorithm of Ben-Or, Kozen, and Reif
\citeyear{BKR} to decide the satisfiability of real closed field
formulas. 
We define the length $\abs{\phi}$ of $\phi$ to be the number of
symbols required to write $\phi$, where we count the length of each
coefficient as $1$. Similarly, we define $\norm{\phi}$ to be the
length of the longest coefficient appearing in $f$, when written in
binary. 

\thm\label{t:complexity1} 
There is a procedure that runs in
space exponential in $\abs{\phi}~\norm{\phi}$ for
deciding, given $\Phih$ and $\Phio$, 
whether a formula $\phi$ of 
$\Lfoev(\Phih,\Phio)$ is satisfiable in an evidential world.  
\ethm
This is essentially the best we can do, since Ben-Or, Kozen, and Reif
\citeyear{BKR} prove that the 
decision problem for real closed fields is complete for
exponential space, and our logic contains the full language of real
closed fields. 

While we assumed that the algorithm takes as input the set of
primitive propositions $\Phih$ and $\Phio$, this does not really
affect the complexity of the algorithm. More precisely, if we are
given a formula $\phi$ in $\Lfoev$ over some set of hypotheses and
observations, we can still decide whether $\phi$ is satisfiable, that
is, whether there are sets $\Phih$ and $\Phio$ of primitive propositions
containing all the primitive propositions in $\phi$
and an evidential world $w$
that satisfies $\phi$.

\thm\label{t:complexity1-alt} 
There is a procedure that runs in space exponential in $\abs{\phi}~\norm{\phi}$ for 
deciding whether there exists sets of primitive 
propositions $\Phih$ and $\Phio$ such that
$\phi\in\Lfoev(\Phih,\Phio)$ and $\phi$ is satisfiable in an
evidential world.
\ethm

The main culprit for the exponential-space complexity is the
theory of real closed fields, which we had to add to the logic to be
able to even write down axiom \axiom{E4} of the axiomatization
$\AX$.%
\footnote{Recall that axiom \axiom{E4} requires existential
quantification. Thus, we can restrict to the sublanguage consisting of
formulas with a single block of existential quantifiers in prefix
position.  The satisfiability problem for this sublanguage can be
shown to be decidable in time exponential in the size of the formula
\cite{r:renegar92}.} 
However, if we are not interested in axiomatizations, but simply in
verifying properties of probabilities and weights of evidence, we can
consider the following propositional (quantifier-free) fragment of our
logic.
As before, we start with sets $\Phih$ and $\Phio$ of hypothesis
and observation primitives, and form the sublanguage $\Lh$ of
hypothesis formulas. 
Basic terms have the form $\Pr^0(\rho)$, $\Pr(\rho)$, and
$\We(\ob,h)$, where $\rho$ is an hypothesis formula, $\ob$ is an
observation, and $h$ is an hypothesis. 
A quantifier-free polynomial term has the form $a_1 t_1+\dots+a_n
t_n$, where each $a_i$ is an integer and each $t_i$ is a product of
basic terms. 
A quantifier-free polynomial inequality formula has the form $p\ge c$,
where $p$ is a quantifier-free polynomial term, and $c$ is an integer.
For instance, a quantifier-free polynomial inequality formula takes
the form $\Pr^0(\rho)+3\We(\ob,h)+5\Pr^0(\rho)\Pr(\rho')\geq 7$.

Let $\Lev(\Phih,\Phio)$ be the language obtained by starting out with
the primitive propositions in $\Phih$ and $\Phio$ and quantifier-free
polynomial inequality formulas, and closing off under conjunction and
negation. 
Since quantifier-free polynomial inequality formulas are polynomial
inequality formulas, $\Lev(\Phih,\Phio)$ is a sublanguage of
$\Lfoev(\Phih,\Phio)$.
The logic $\Lev(\Phih,\Phio)$ is sufficiently expressive to express
many properties of interest; for instance, it can certainly express
the general connection between priors, posteriors, and evidence
captured by axiom \axiom{E3}, as well as specific relationships
between prior probability and posterior probability through the weight
of evidence of a particular observation, as in
Example~\ref{x:coins2}. 
Reasoning about the propositional fragment of our logic
$\Lev(\Phih,\Phio)$ is easier than the full language.%
\footnote{In a preliminary version of this paper \cite{r:halpern03b},
  we examined the quantifier-free fragment of $\Lfoev(\Phih,\Phio)$
  that uses only linear inequality formulas, of the form $a_1
  t_1+\dots+a_n t_n\ge c$, where each $t_i$ is a basic term. 
 We claimed that the problem of deciding, given $\Phih$
 and $\Phio$, whether a formula $\phi$ of this fragment is satisfiable
 in an evidential world is NP-complete. 
We further claimed that this result followed from a small-model theorem: if
 $\phi$ is satisfiable, then it is satisfiable in an evidential world over
 a small  number of hypotheses and observations. 
While this small-model theorem is true, our argument that the
satisfiability problem is in NP also implicitly assumed that the 
numbers associated with the probability  measure 
and the evidence space in the evidential world were small. 
But this is not true in general.  Even though the formula $\phi$
involves only  
 linear inequality formulas, every evidential world satisfies 
 axiom \axiom{E3}. 
This constraint enables us to write formulas for which there exist no
models where the probabilities and weights of evidence are rational.
For example, consider the formula
\[ \Pr^0(h_1) = \We(\ob_1,h_1)  \land
   \Pr^0(h_2) = 1-\Pr^0(h_1) \land \Pr(h_1) = 1/2 \land \We(\ob_1,h_2)
   = 1/4 \] 
  Any evidential world satisfying the formula must satisfy
\[ \Pr^0(h_1) = \We(\ob_1,h_1) = -1/8 (1-\sqrt{17})\]
which is irrational. 
The exact complexity of this fragment remains open.  We can use our
techniques to show that it is in PSPACE, but we have no matching lower
bound.  (In particular, it may indeed be in NP.)
We re-examine this fragment of the logic in
Section~\ref{s:normalized}, under a different interpretation of
weights of evidence.}

\thm\label{t:complexity2} 
There is a procedure that runs in space
polynomial in $\abs{\phi}~\norm{\phi}$ for deciding, given $\Phih$ and
$\Phio$, whether a formula $\phi$ of $\Lev(\Phih,\Phio)$ is
satisfiable in an evidential world.  
\ethm

Theorem~\ref{t:complexity2} relies on Canny's \citeyear{r:canny88}
procedure for deciding the validity of quantifier-free formulas in the
theory of real closed fields.
As in the general case, the complexity is unaffected by whether or not 
the decision problem takes as input the sets $\Phih$ and $\Phio$ of
primitive propositions. 

\thm\label{t:complexity2-alt}
There is a procedure that runs in space polynomial in
$\abs{\phi}~\norm{\phi}$ for deciding whether there exists sets of
primitive  propositions $\Phih$ and $\Phio$ such that 
$\phi\in\Lev(\Phih,\Phio)$ and $\phi$ is satisfiable in an
evidential world.
\ethm

\section{Normalized Versus Unnormalized Likelihoods}\label{s:normalized}

The weight of evidence we used throughout this paper is a
generalization of the log-likelihood ratio advocated by Good
\citeyear{r:good50,r:good60}. 
As we pointed out earlier,
this measure of confirmation is essentially
a normalized likelihood: the likelihood of an observation given
a particular hypothesis is normalized by the sum of all the
likelihoods of that observation, for all possible hypotheses. 
What would change if we were to take the 
(unnormalized)
likelihoods $\mu_h$
themselves as weight of evidence? 
Some things would simplify.
For example, \axiom{WF2} is a consequence of normalization, as is the
corresponding 
axiom \axiom{E4}, which is the only axiom that requires
quantification.

The main argument for normalizing likelihood is the same
as that for normalizing probability measures. 
Just like probability, when using normalized likelihood, the weight of
evidence is always between 0 and 1, and provides an absolute scale
against which to judge all reports of evidence. 
The impact here is psychological---it permits one to use the same 
rules of thumb in all situations, since the numbers obtained are independent
from the context of their use. 
Thus, for instance, a weight of evidence of 0.95 in one situation
corresponds to the ``same amount'' of evidence as a weight of evidence
of 0.95 in a different situation;
any acceptable decision based on this weight of evidence in the first
situation ought to be acceptable in the other situation as well. 
The importance of having such a uniform scale depends, of course, on the
intended applications.

For the sake of completeness, we now describe the changes to our
framework required to use unnormalized likelihoods as a weight of
evidence. Define $w^u_\cE(\ob,h)=\mu_h(\ob)$. 

First, note that we can update a prior probability $\mu_0$ via a set of
likelihood functions $\mu_h$ using a form of Dempster's Rule of
Combination.  
More precisely, we can define $\mu_0\oplus w^u_\cE(\ob,\cdot)$ to be the
probability measure defined by \[(\mu_0\oplus w^u_{\cE}(\ob,\cdot))(h)
= \frac{\mu_0(h)\mu_h(\ob)}{\sum_{h'\in\cH}\mu_0(h')\mu_{h'}(\ob)}.\]

The logic we introduced in Section~\ref{s:logic} applies just as well
to this new interpretation of weights of evidence. The syntax remains
unchanged, the models remain evidential worlds, and the semantics of
formulas simply take the new interpretation of weight of evidence into
account. 
In particular, the assignment $\val{p}{w,v}$ now uses the above
definition of $w^u_{\cE}$, and becomes
\begin{eqnarray*}
\val{\Pr(\rho)}{w,v} & = &
    (\mu\oplus w^u_{\cE}(\ob,\cdot))(\intension{\rho})\\ 
\val{\We(\ob',h')}{w,v} & = & w^u_{\cE}(\ob',h').
\end{eqnarray*}

The axiomatization of this new logic is slightly different
and somewhat simpler than the one in Section~\ref{s:logic}. 
In particular, \axiom{E1} and \axiom{E2}, which say that $\We(\ob,h)$
acts as a probability measure for each fixed $\ob$, are replaced by
axioms that say that $\We(\ob,h)$ acts as a probability measure for
each fixed $h$: 
\begin{axiomlist}
\item[\axiom{E1$'$}.] $\We(\ob,h)\geq 0$.
\item[\axiom{E2$'$}.] $\We(\ob_1,h)+\dots+\We(\ob_{\no},h)=1$.
\end{axiomlist}
Axiom \axiom{E3} is unchanged, since $w^u_\cE$ is updated in 
essentially the same way as $w_\cE$.
Axiom \axiom{E4} becomes unnecessary.

What about the complexity of the decision procedure? As in
Section~\ref{s:decision}, the complexity of the decision problem for
the full logic $\Lfoev(\Phih,\Phio)$ remains dominated by the
complexity of reasoning in real closed fields. Of course, now, we can
express the full axiomatization for the unnormalized likelihood
interpretation of weight of evidence in the $\Lev(\Phih,\Phio)$
fragment, which can be decided in polynomial space. 
A further advantage of the unnormalized likelihood interpretation of weight
of evidence, however,  is that it leads to a useful fragment of
$\Lev(\Phih,\Phio)$  that is perhaps easier to decide.  

Suppose that we are interested in reasoning exclusively about weights of
evidence, with no prior or posterior probability. This is the kind of
reasoning that actually underlies many computer science applications
involving randomized algorithms \cite{r:halpern05c}.
As before, we start with sets $\Phih$ and $\Phio$ of hypothesis and
observation primitives, and form the sublanguage $\Lh$ of hypothesis
formulas. 
A quantifier-free linear term has the form $a_1
\We(\ob^1,h^1)+\dots+a_n \We(\ob^n,h^n)$, where each $a_i$ is an
integer, each $\ob^i$ is an observation, and each $h^i$ is an
hypothesis. 
A quantifier-free linear inequality formula has the form $p\ge c$,
where $p$ is a quantifier-free linear term and $c$ is an integer.
For example, $\We(\ob',h)+3\We(\ob,h)\geq 7$ is a quantifier-free linear
inequality formula.

Let $\Lw(\Phih,\Phio)$ be the language obtained by starting out with
the primitive propositions in $\Phih$ and $\Phio$ and quantifier-free
linear inequality formulas, and closing off under conjunction and
negation. Since quantifier-free linear inequality formulas are
polynomial inequality formulas, $\Lw(\Phih,\Phio)$ is a sublanguage
of $\Lfoev(\Phih,\Phio)$.
Reasoning about $\Lw(\Phih,\Phio)$ is easier than the full language,
and possibly easier than the $\Lev(\Phih,\Phio)$ fragment. 

\thm\label{t:complexity3}
The problem of deciding, given $\Phih$ and $\Phio$, whether a formula
$\phi$ of $\Lw(\Phih,\Phio)$ is satisfiable in an evidential world
is NP-complete. 
\ethm

As in the general case, the complexity is unaffected by whether or not 
the decision problem takes as input the sets $\Phih$ and $\Phio$ of
primitive propositions. 

\thm\label{t:complexity3-alt}
The problem of deciding, for a formula $\phi$, whether there exists
sets of primitive propositions $\Phih$ and $\Phio$ such that
$\phi\in\Lw(\Phih,\Phio)$ and $\phi$ is satisfiable in an
evidential world is NP-complete.
\ethm

\section{Evidence in Dynamic Systems}\label{s:dynamic}

The evidential worlds we have considered until now are
essentially static, in that they model only the situation where a
single observation is made. 
Considering such static worlds lets us
focus on the relationship between the prior and posterior
probabilities on hypotheses and the weight of evidence of a single
observation.  In a related paper \cite{r:halpern05c}, we consider
evidence in the context of randomized algorithms; we use evidence to
characterize the information provided by, for example, a randomized
algorithm for primality when it says that a number is prime.  The
framework in that work is dynamic; sequences of
observations are made over time.
In this section, we extend our logic to reason about the evidence of
sequences of observations, using the approach to combining evidence
described in Section~\ref{s:evidence}. 

There are subtleties involved in trying to find an appropriate logic for
reasoning about situations like that in Example~\ref{x:coinsseq}.
The most important one is the relationship
between observations and time. 
By way of illustration, consider the following example. Bob is
expecting an email from Alice stating where a rendezvous is to take
place. Calm under pressure, Bob is reading while he waits. We assume
that Bob is not concerned with the time. For the
purposes of this  example, one of three things can occur at any given
point in time: 
\begin{enumerate}
\item Bob does not check if he has received email;
\item Bob checks if he has received email, and notices he has not
received an email from Alice;
\item Bob checks if he has received email, and notices he has received 
an email from Alice.
\end{enumerate}
How is his view of the world affected by these events? In (1), it
should be clear that, all things being equal, Bob's view of the world
does not change: no observation is made. Contrast this with (2) and
(3). In (2), Bob does make an observation, namely that he has not yet
received Alice's email. The fact that he checks indicates that he
wants to observe a result. In (3), he also makes an observation,
namely that he received an email from Alice. In both of these cases,
the check yields an observation, that he can use to update his view of
the world. In case (2), he essentially observed that nothing
happened, but we emphasize again that this is an observation, to be
distinguished from the case where Bob does not even check whether
email has arrived, 
and should be explicit in the set $\cO$ in the evidence space.

This discussion motivates the models that we use in this
section. 
We characterize an agent's state by the observations that she has made,
including possibly the ``nothing happened'' observation.  Although we do not
explicitly model time, it is easy to incorporate time in our framework,
since the agent can observe times or clock ticks.
The models in this section are admittedly simple, but they already
highlight the issues involved in reasoning about evidence in dynamic
systems. As long as agents do not forget observations, there is no
loss of generality in associating an agent's state with a sequence of
observations. We do, however, make the simplifying assumption that the
same evidence space is used for all the observations in a sequence. In
other words, we assume that the evidence space is fixed for the
evolution of the system. 
In many situations of interest, the external world changes.  The
possible observations may depend on the state of the world, as may the
likelihood functions.  
There are no intrinsic difficulties in extending the model to handle 
state changes, but the additional details would only obscure the
presentation.   

In some ways, considering a dynamic setting simplifies things.  Rather than
talking about the prior and posterior probability using different
operators, we need only a single probability operator that represents
the probability of an hypothesis at the current time.  To express the
analogue of axiom \axiom{E3} in this logic, we need to be able to talk
about the probability at the next time step. This can be done by
adding the ``next-time'' operator $\Circ$ to the logic, where
$\Circ\phi$ holds at the current time if $\phi$ holds at the next time
step.%
\footnote{Following the discussion above, time steps are associated with
new  observations. Thus, $\Circ\phi$ means that $\phi$ is true at the next
time step, that is, after the next observation. This simplifies the
presentation of the logic.}
We further extend the logic to talk about the weight of
evidence of a sequence of observations.  

We define the logic $\Lfoevdyn$ as follows. As in
Section~\ref{s:logic}, we start with a set of primitive propositions
$\Phih$ and $\Phio$, respectively representing the hypotheses and the
observations. Again, let $\Lh(\Phih)$ be the propositional sublanguage 
of hypotheses formulas obtained by taking primitive propositions
in $\Phih$ and closing off under negation and conjunction; we use
$\rho$ to range over formulas of that sublanguage. 

A basic term now has the form $\Pr(\rho)$ or $\We(\sequ{\ob},h)$, where
$\rho$ is an hypothesis formula, $\sequ{\ob}=\<\ob^1,\dots,\ob^k\>$
is a nonempty sequence of observations, and $h$ is an hypothesis.  
If $\sequ{\ob}=\<\ob^1\>$, 
we write $\We(\ob_1,h)$ rather than $\We(\<\ob^1\>,h)$.
As before, a polynomial term has the 
form $t_1+\dots+t_n$, where each term $t_i$ is a product of integers,
basic terms, and variables (which intuitively range over the reals). A
polynomial inequality formula has the form $p\ge c$, where $p$ is a
polynomial term and $c$ is an integer. Let $\Lfoevdyn(\Phih,\Phio)$
be the language obtained by starting out with the primitive
propositions in $\Phih$ and $\Phio$ and polynomial inequality
formulas, and closing off under conjunction, negation, first-order
quantification, and application of the $\Circ$ operator.
We use the same abbreviations as in Section~\ref{s:logic}. 

The semantics of this logic now involves models that have dynamic
behavior. 
Rather than just considering individual worlds, we now consider
sequences of worlds, which 
we call {\em runs}, representing the
evolution of the system over time. 
A model is now an infinite run, where a run
describes a possible dynamic evolution of the system. As before, a run
records the observations being made and the hypothesis that is true
for the run, as well as a probability distribution describing the
prior probability of the hypothesis at the initial state of the run,
and an evidence space $\cE^*$ over $\Phih$ and $\Phio^*$ to interpret
$\We$.  We define an \emph{evidential run} $r$ to be a map from the
natural numbers (representing time) to histories of the system up to
that time. 
A history at time $m$ records the relevant information about the
run---the hypothesis that is true, the prior probability on the
hypotheses, and the evidence space $\cE^*$---and the observations that have
been made up to time $m$.
Hence, a history has the form
$\<(h,\mu,\cE^*),\ob^1,\dots,\ob^k\>$.  We assume that
$r(0)=\<(h,\mu,\cE^*)\>$ for some $h$, $\mu$, and $\cE^*$, while
$r(m)=\<(h,\mu,\cE^*),\ob^1,\dots,\ob^m\>$ for $m>0$. We define a
point of the run to be a pair $(r,m)$ consisting of a run $r$ and
time $m$.

We associate with each propositional formula $\rho$ in $\Lh(\Phih)$ a
set $\intension{\rho}$ of hypotheses, just as we did in
Section~\ref{s:logic}.

In order to ascribe a semantics to first-order formulas that may
contain variables, we need a valuation $v$ that assigns a real number
to every variable.  Given a valuation $v$, an evidential run $r$, and
a point $(r,m)$,  where 
$r(m)=\<(h,\mu,\cE^*),\ob^1,\dots,\ob^m\>$, 
we can assign to a
polynomial term $p$ a real number $\val{p}{r,m,v}$ 
using essentially the same approach as in Section~\ref{s:logic}:
\begin{eqnarray*}
\val{x}{r,m,v} & = & v(x)\\
\val{a}{r,m,v} & = & a\\
\val{\Pr(\rho)}{r,m,v} & = &
    (\mu\oplus w_{\cE^*}(\<\ob^1,\dots,\ob^m\>,\cdot)))(\intension{\rho})\\ 
& & \quad\mbox{where $r(m)=\<(h,\mu,\cE^*),\ob^1,\dots,\ob^m\>$}\\
\val{\We(\sequ{\ob},h')}{r,m,v} & = &
w_{\cE^*}(\sequ{\ob},h')\\
& & \quad\mbox{where $r(m)=\<(h,\mu,\cE^*),\ob^1,\dots,\ob^m\>$}\\
\val{t_1 t_2}{r,m,v} & = & \val{t_1 }{r,m,v} \times
\val{t_2}{r,m,v}\\
\val{p_1 + p_2}{r,m,,v} & = & \val{p_1}{r,m,v} + \val{p_2}{r,m,v}.
\end{eqnarray*}

We define what it means for a formula $\phi$ to be true (or satisfied)
at a point $(r,m)$ of an evidential run $r$
under valuation $v$, written $(r,m,v)\sat\phi$, 
using essentially the same approach as in Section~\ref{s:logic}:
\begin{itemize}
\item[]$(r,m,v)\sat h$ if $r(m)=\<(h,\mu,\cE^*),\dots\>$
\item[]$(r,m,v)\sat\ob$ if $r(m)=\<(h,\mu,\cE^*),\dots,\ob\>$
\item[]$(r,m,v)\sat\neg\phi$ if $(r,m,v)\not\sat\phi$
\item[]$(r,m,v)\sat\phi\land\psi$ if $(r,m,v)\sat\phi$ and
$(r,m,v)\sat\psi$
\item[]$(r,m,v)\sat p \ge c$ if $\val{p}{r,m,v} \ge c$
\item[]$(r,m,v)\sat \Circ\phi$ if $(r,m+1,v)\sat\phi$
\item[]$(r,m,v)\sat\forall x\phi$ if $(r,m,v')\sat\phi$ for
all valuations $v'$ that agree with $v$ on all variables but $x$. 
\end{itemize}
If $(r,m,v)\sat\phi$ is true for all $v$, we simply write
$(r,m)\sat\phi$.  If $(r,m)\sat\phi$ for all points $(r,m)$ of $r$,
then we write $r\sat\phi$ and say that $\phi$ is valid in
$r$. Finally, if $r\sat\phi$ for all evidential runs $r$, we write
$\sat\phi$ and say that $\phi$ is valid.

It is straightforward to axiomatize this new logic. The
axiomatization shows 
that
we can capture the combination of evidence
directly in the logic, a pleasant property.
Most of the axioms
from Section~\ref{s:logic} carry over immediately. Let the
axiomatization $\AXdyn$ consists of the following axioms and inference
rules: first-order
reasoning (\axiom{Taut}, \axiom{MP}), reasoning about polynomial
inequalities (\axiom{RCF}), reasoning about hypotheses and
observations (\axiom{H1},\axiom{H2},\axiom{O1},\axiom{O2}), reasoning
about probabilities (\axiom{Po1}--\axiom{4} only, since we do not have
$\Pr^0$ in the language), and reasoning about weights of evidence
(\axiom{E1}, \axiom{E2}, \axiom{E4}), as well as new axioms we now
present. 

Basically, the only axiom that needs replacing is \axiom{E3}, which links
prior and posterior probabilities, since this now needs to be
expressed using the $\Circ$ operator. Moreover, we need an axiom to
relate the weight of evidence of a sequence of observation to the
weight of evidence of the individual observations,
as given by Equation (\ref{e:decomposition}).
\begin{axiomlist}
\item[\axiom{E5}.]
$\begin{prog}
 \ob\rimp
 \forall x(\begin{prog}
                  \Circ(\Pr(h)=x)\rimp\\
                  \quad \Pr(h)\We(\ob,h)=
                   x \Pr(h_1)\We(\ob,h_1) + \dots +
                   x \Pr(h_{\nh})\We(\ob,h_{\nh})).
                  \end{prog}
 \end{prog}$
\item[\axiom{E6}.] 
$\begin{prog}
\We(\ob^1,h)\cdots\We(\ob^k,h) = 
\begin{prog}
\We(\<\ob^1,\dots,\ob^k\>,h)\We(\ob^1,h_1)\cdots\We(\ob^k,h_1)
+ \dots + \\ \quad
\We(\<\ob^1,\dots,\ob^k\>,h)\We(\ob^1,h_{\nh})\cdots\We(\ob^k,h_{\nh}).
\end{prog} 
\end{prog} 
$
\end{axiomlist}

To get a complete axiomatization, we also need axioms 
and inference rules
that capture the
properties of the 
temporal operator $\Circ$. 
\begin{axiomlist}
\item[\axiom{T1}.] $\Circ\phi\land\Circ(\phi\rimp\psi)\rimp\Circ\psi$.
\item[\axiom{T2}.] $\Circ\neg\phi\riff\neg\Circ\phi$.
\item[\axiom{T3}.] From $\phi$ infer $\Circ\phi$.
\end{axiomlist}
Finally, we need axioms to say that the truth of hypotheses as well as
the value of polynomial terms not containing occurrences of $\Pr$ is
time-independent:
\begin{axiomlist}
\item[\axiom{T4}.] $\Circ\rho\riff \rho$.
\item[\axiom{T5}.] $\Circ(p\ge c) \riff p\ge c$ if $p$ does not
contain an occurrence of $\Pr$. 
\item[\axiom{T6}.] $\Circ(\forall x\phi)\riff\forall x(\Circ\phi)$.
\end{axiomlist}

\thm\label{t:sound-complete-dyn}
$\AXdyn$ is a sound and complete axiomatization for
$\Lfoevdyn(\Phih,\Phio)$ with respect to evidential runs.
\ethm

\section{Conclusion}\label{s:conclusion}

In the literature, reasoning about the effect of observations is
typically done in a context where we have a prior probability on a set 
of hypotheses which we can condition on the observations made to obtain a
new probability on the hypotheses that reflects the effect of the
observations. In this paper, we have presented a logic of evidence
that lets us reason about the weight of evidence of observations,
independently of any prior probability on the hypotheses. The logic is 
expressive enough to capture in a logical form the relationship
between a prior probability on hypotheses, the weight of evidence of
observations, and the result posterior probability on hypotheses. But
we can also capture reasoning that does not involve prior
probabilities. 

While the logic is essentially propositional, obtaining a sound and
complete axiomatization seems to require quantification over the
reals. This adds to the complexity of the logic---the decision problem 
for the full logic is in exponential space. However, an interesting
and potentially useful fragment, the propositional fragment, is
decidable in polynomial space. 

\paragraph{Acknowledgments.}

A preliminary version of this paper appeared in the \emph{Proceedings
of the Nineteenth Conference on Uncertainty in Artificial
Intelligence}, pp. 297--304, 2003. 
This work was mainly done while the second author was at Cornell
University. 
We thank Dexter Kozen 
and Nimrod Megiddo 
for useful discussions.  
Special thanks to Manfred Jaeger for his careful reading of the paper 
and subsequent comments.
Manfred found the bug in our proof that the satisfiability problem for
the quantifier-free fragment of $\Lfoev(\Phih,\Phio)$ that uses only
linear inequality formulas is NP-complete.
His comments also led us to discuss the issue of normalization.
We also thank the reviewers, whose comments greatly improved the
paper. 
This work was supported in part by NSF under grants
CTC-0208535, ITR-0325453, and IIS-0534064, by ONR under grant
N00014-01-10-511, by the DoD Multidisciplinary University Research
Initiative (MURI) program administered by the ONR under
grants N00014-01-1-0795 and N00014-04-1-0725, and by AFOSR under grant
F49620-02-1-0101.

\appendix

\section{Proofs}

\opro{p:induces-same}
For all $\ob$, we have $w_{\cE}(\ob,h_i)\geq
w_{\cE}(\ob,h_{3-i})$ if and  only if 
$l(\ob,h_i)\ge l(ob,h_{3-i})$, for $i = 1, 2$, and for all $h$, $\ob$,
and $\ob'$, we have  $w_{\cE}(\ob,h)\geq w_{\cE}(\ob',h)$ if and only
if $l(\ob,h)\geq l(\ob',h)$.  
\eopro
\prf
Let $\ob$ be an arbitrary observation. The result follows from the
following argument:
\begin{align*}
& w_{\cE}(\ob,h_i)\geq w_{\cE}(\ob,h_{3-i})\\
& \qquad\mbox{iff }
\mu_{h_i}(\ob)/(\mu_{h_i}(\ob)+\mu_{h_{3-i}}(\ob))\ge\mu_{h_{3-i}}(\ob)/ 
(\mu_{h_i}(\ob)+\mu_{h_{3-i}}(\ob))\\  
& \qquad\mbox{iff }
\mu_{h_i}(\ob)\mu_{h_i}(\ob)\ge\mu_{h_{3-i}}(\ob)\mu_{h_{3-i}}(\ob)\\
& \qquad\mbox{iff }
\mu_{h_i}(\ob)/\mu_{h_{3-i}}(\ob)\ge\mu_{h_{3-i}}(\ob)/\mu_{h_i}(\ob)\\
& \qquad\mbox{iff } l(\ob,h_i) \ge l(ob,h_{3-i}).
\end{align*}
A similar argument establishes the result for hypotheses.
\eprf

\othm{t:characterization}
Let $\cH=\{h_1,\dots,h_m\}$ and $\cO=\{\ob_1,\dots,\ob_n\}$, and let
$f$ be a real-valued function with domain $\cO\times\cH$ such 
that $f(\ob,h)\in[0,1]$. Then there exists an evidence space
$\cE=(\cH,\cO,\mu_{h_1},\dots,\mu_{h_m})$ such that $f=w_{\cE}$ if
and only if $f$ satisfies the following properties:
\begin{axiomlist}
\item[\rm\axiom{WF1}.] For every $\ob\in\cO$, $f(\ob,\cdot)$ is a
probability measure on $\cH$.
\item[\rm\axiom{WF2}.] There exists $x_1,\dots,x_n > 0$ such that, for all
$h\in\cH$, $\sum_{i=1}^{n}f(\ob_i,h)x_i=1$.
\end{axiomlist}
\eothm

\prf
$(\Rightarrow)$ Assume that $f=w_{\cE}$ for some evidence space
$\cE=(\cH,\cO,\mu_{h_1},\dots,\mu_{h_m})$. It is routine to verify
\axiom{WF1}, that for a fixed $\ob\in\cO$, $w_{\cE}(\ob,\cdot)$ is a
probability measure on $\cH$. To verify \axiom{WF2}, note that we can
simply take $x_i = \sum_{h'\in\cH} \mu_{h'}(\ob_i)$. 

$(\Leftarrow)$ Let $f$ be a function from $\cO\times\cH$ to $[0,1]$
that satisfies \axiom{WF1} and \axiom{WF2}. Let
$x^*_1,\dots,x^*_{\nh}$ be the positive reals guaranteed by
\axiom{WF2}. It is straightforward to verify that taking
$\mu_h(\ob_i) = f(\ob_i,h)/x^*_i$ for each $h\in\cH$ yields an
evidence space $\cE$ such that $f=w_{\cE}$. 
\eprf

The following lemmas are useful to prove the completeness of the
axiomatizations in this paper. 
These results depend on the soundness of the axiomatization $\AX$.
\lem\label{l:soundness}
$\AX$ is a sound axiomatization for the logic
$\Lfoev(\Phih,\Phio)$ with respect to evidential worlds.
\elem
\prf
It is easy to see that each axiom is valid in evidential worlds.
\eprf

\lem\label{l:intension}
For all hypothesis formulas $\rho$,  $\rho\riff h_1\lor\dots\lor h_k$
is provable in  $\AX$, when $\intension{\rho}=\{h_1,\dots,h_k\}$.  
\elem
\prf
Using \axiom{Taut}, we can show that $\rho$ is provably
equivalent to a formula $\rho'$ in disjunctive normal form. Moreover,
by axiom \axiom{H2}, we can assume without loss of generality that each 
of the disjuncts in $\rho'$ consists of a single hypothesis. Thus, $\rho$ is
$h_1\lor\dots\lor h_k$. 
An easy induction on structure shows that for an hypothesis formula
$\rho$ and evidential world $w$, we have that
$w \sat \rho$ iff $w \sat h$ for some $h \in \intension{\rho}$.
Moreover, it follows immediately from the soundness of the
axiomatization (Lemma~\ref{l:soundness}) that $\rho \riff h_1 \lor
\ldots \lor h_k$ is provable iff for all evidential worlds $w$,  $w
\sat \rho$ iff $w \sat h_i$ for some $i \in \{1,\ldots, k\}$.  Thus,
$\rho \riff h_1 \lor \ldots \lor h_k$ is provable iff
$\intension{\rho} = \{h_1, \ldots, h_k\}$. 
\eprf

An easy consequence of Lemma~\ref{l:intension} is that $\rho_1$ is
provably equivalent to $\rho_2$ if and only if
$\intension{\rho_1}=\intension{\rho_2}$.

\lem\label{l:decompose}
Let $\rho$ be an hypothesis formula. The formulas
\begin{itemize}
\item[] $\Pr(\rho)=\sum\limits_{h\in\intension{\rho}} \Pr(h)$ and
\item[] $\Pr^0(\rho)=\sum\limits_{h\in\intension{\rho}} \Pr^0(h)$
\end{itemize}
are provable in $\AX$. 
\elem
\prf 
Let $\Phih=\{h_1,\dots,h_{\nh}\}$ and $\Phio=\{\ob_1,\dots,\ob_{\no}\}$. 
We prove the result for $\Pr$. 
We proceed by induction on the size of $\intension{\rho}$. For the
base case, assume that $|\intension{\rho}|=0$. By
Lemma~\ref{l:intension}, this implies that $\rho$ is provably
equivalent to $\falsep$. By \axiom{Po4}, $\Pr(\rho)=\Pr(\falsep)$, and
it is easy to check that $\Pr(\falsep)=0$ is provable using
\axiom{Po1}, \axiom{Po3}, and \axiom{Po4}, thus $\Pr(\rho)=0$, as
required. If $|\intension{\rho}|=n+1>0$, then
$\intension{\rho}=\{h_{i_1},\dots,h_{i_{n+1}}\}$, and by
Lemma~\ref{l:intension}, $\rho$ is provably equivalent to
$h_{i_1}\lor\dots\lor h_{i_{n+1}}$. By \axiom{Po4},
$\Pr(\rho)=\Pr(\rho\land h_{i_{n+1}})+\Pr(\rho\land\neg
h_{i_{n+1}})$. It is easy to check that $\rho\land h_{i_{n+1}}$ is
provably equivalent to $h_{i_{n+1}}$ (using \axiom{H2}), and similarly
$\rho\land\neg h_{i_{n+1}}$ is provably equivalent to
$h_{i_1}\lor\dots\lor h_{i_n}$. Thus,
$\Pr(\rho)=\Pr(h_{i_{n+1}})+\Pr(h_{i_1}\lor\dots\lor h_{i_n})$ is
provable. Since $|\intension{h_{i_1}\lor\dots\lor h_{i_n}}|=n$, by the
induction hypothesis, $\Pr(h_{i_1}\lor\dots\lor
h_{i_n})=\sum_{h\in\{h_{i_1},\dots,h_{i_n}\}}\Pr(h) =
\sum_{h\in\intension{\rho}-\{h_{i_{n+1}}\}}\Pr(h)$. Thus,
$\Pr(\rho)=\Pr(h_{i_{n+1}})+\sum_{h\in\intension{\rho}-\{h_{i_{n+1}}\}}\Pr(h)$, 
that is, $\Pr(\rho)=\sum_{h\in\intension{\rho}}\Pr(h)$, as required.

The same argument applies \emph{mutatis mutandis} for $\Pr^0$, using
axioms \axiom{Pr1}--\axiom{4} instead of \axiom{Po1}--\axiom{4}.  
\eprf

\othm{t:sound-complete} 
$\AX$ is a sound and complete axiomatization
for the logic with respect to evidential worlds.
\eothm 

\prf 
Soundness was established in Lemma~\ref{l:soundness}. 
To prove completeness, recall the following definitions.
A formula $\phi$ is \emph{consistent} with the axiom system $\AX$ if
$\neg\phi$ is not provable from $\AX$. 
To prove completeness, it is sufficient to show that if $\phi$ is
consistent, then it is satisfiable, that is, there exists an
evidential world $w$ and valuation $v$ such that
$(w,v)\sat\phi$. 

As in the body of the paper, let $\Phih=\{h_1,\dots,h_{\nh}\}$ and
$\Phio=\{\ob_1,\dots,\ob_{\no}\}$. 
Let $\phi$ be a consistent formula. By way of contradiction, assume
that
$\phi$ is unsatisfiable. We reduce the formula $\phi$ to an
equivalent formula in the language of real closed fields.  Let
$u_1,\dots,u_{\nh}$, $v_1,\dots,v_{\no}$, $x_1,\dots,x_{\nh}$,
$y_1,\dots,y_{\no}$, and $z^1_1,\dots,z^1_{\nh},\dots,z^{\no}_1,\dots,z^{\no}_{\nh}$
be new variables, where, intuitively,
\begin{itemize}
\item $u_i$ gets value $1$ if hypothesis $h_i$ holds, $0$ otherwise;
\item $v_i$ gets value $1$ if observation $\ob_i$ holds, $0$
otherwise;
\item $x_i$ represents $\Pr^0(h_i)$;
\item $y_i$ represents $\Pr(h_i)$;
\item $z_{i,j}$ represents $\We(\ob_i,h_j)$.
\end{itemize}
Let $\mathbf{v}$ represent that list of new variables. Consider the
following formulas. 
Let $\phihyp$ be the formula saying that exactly one hypothesis
holds:
\[(u_1=0\lor u_1=1)\land\dots\land(u_{\nh}=0\lor
u_{\nh}=1)\land u_1+\dots+u_{\nh}=1.\]
Similarly, let $\phiobs$ be the formula saying that exactly one
observation holds:
\[(v_1=0\lor v_1=1)\land\dots\land(v_{\no}=0\lor
v_{\nh}=1)\land v_1+\dots+v_{\nh}=1.\]
Let $\phiprior$ be the formula that expresses that $\Pr^0$ is a
probability measure:
\[\phiprior=x_1\ge 0\land\dots\land x_{\nh}\ge 0 \land
x_1+\dots+x_{\nh}=1.\]
Similarly, let $\phipost$ be the formula that expresses that $\Pr$ is
a probability measure:
\[\phipost=y_1\ge 0\land\dots\land y_{\nh}\ge 0\land
y_1+\dots+y_{\nh}=1.\]
Finally, we need formulas saying that $\We$ is a weight of evidence
function. The formula $\phiwprob$ simply says that $\We$ satisfies
\axiom{WF1}, that is, it acts as a probability measure for a fixed
observation:
\begin{align*}
  & z_{1,1}\ge 0\land\dots\land z_{1,\nh}\ge 0\land z_{\no,1}\ge 0\land\dots\land z_{\no,\nh}\ge 0\land \\
  & z_{1,1}+\dots+z_{1,\nh}=1 \land \dots \land z_{\no,1}+\dots+z_{\no,\nh}=1.
\end{align*}
The formula $\phiwf$ says that $\We$ satisfies \axiom{WF2}:
\begin{align*}
   & \exists w_1,\dots,w_{\no}(\begin{prog}
        w_1> 0\land\dots\land w_{\no}> 0 \land
                          z_{1,1} w_1+\dots+ z_{\no,1} w_{\no}=1 \land\\
                                   \dots \land 
                                   z_{1,\nh} w_1+\dots+ z_{\no,\nh} w_{\no}=1)
                                   \end{prog}\\
   & \quad\mbox{where $w_1,\dots,w_{\no}$ are new variables.}
\end{align*}
Finally, the formula $\phiwup$ captures the fact that weights of
evidence can be viewed as updating a prior probability into a
posterior probability, via Dempster's Rule of Combination:
\begin{align*}
  & (v_1=1 \rimp (\begin{prog}
      x_1 z_{1,1} = y_1 x_1 z_{1,1} + \dots + y_1 x_{\nh} z_{1,\nh}\land\\
                 \dots\land
                 x_{\nh} z_{1,\nh} = y_{\nh} x_1 z_{1,1} + \dots + y_{\nh} x_{\nh} z_{1,\nh}))\land
                 \end{prog}\\
  & \dots\land \\
  & (v_{\no}=1 \rimp (\begin{prog}
                     x_1 z_{\no,1} = y_1 x_1 z_{\no,1} + \dots + y_1 x_{\nh} z_{\no,\nh}\land\\
                     \dots\land
                     x_{\nh} z_{\no,\nh} = y_{\nh} x_1 z_{\no,1} + \dots y_{\nh} x_{\nh} z_{\no,nh})).
           \end{prog}
\end{align*}

Let $\hat{\phi}$ be the formula in the language of real closed fields
obtained from $\phi$ by replacing each occurrence of the primitive
proposition $h_i$ by $u_i=1$, each occurrence of $\ob_i$ by $v_i=1$,
each occurrence of $\Pr^0(\rho)$ by $\sum_{h_i\in\intension{\rho}}x_i$,
each occurrence of $\Pr(\rho)$ by $\sum_{h_i\in\intension{\rho}}y_i$,
each occurrence of $\We(\ob_i,h_j)$ by $z_{i,j}$, and each occurrence of an integer
coefficient $k$ by $1+\dots+1$ ($k$ times).  
Finally, let $\phi'$ be the formula
$\exists\mathbf{v}(\phihyp\land\phiobs\land\phiprior\land\phipost\land\phiwprob\land\phiwf\land\phiwup\land\hat{\phi})$.

It is easy to see that if $\phi$ is unsatisfiable over evidential
worlds, then $\phi'$ is false when interpreted over the real
numbers. Therefore, 
$\neg\phi'$ must be a 
formula valid in real closed fields, and
hence an instance of \axiom{RCF}. Thus, $\neg\phi'$ is provable. It is
straightforward to show, using Lemma~\ref{l:decompose}, that $\neg\phi$
itself is provable, contradicting the fact that $\phi$ is
consistent. Thus, $\phi$ must be satisfiable, establishing
completeness. 
\eprf

As we mentioned at the beginning of Section~\ref{s:decision}, $\Lfoev$ 
is not monotone with respect to validity: axiom \axiom{H1} depends on
the set of hypotheses and observations, and will in general no longer 
be valid if the set is changed. The same is true for \axiom{O1},
\axiom{E3}, and \axiom{E4}. We do, however, have a form of monotonicity
with respect to satisfiability, as the following lemma shows.

\lem\label{l:smallsets}
Given $\Phih$ and $\Phio$, let $\phi$ be a formula of
$\Lfoev(\Phih,\Phio)$, and let $\cH\subseteq\Phih$ and
$\cO\subseteq\Phio$ be the hypotheses and observations that occur in
$\phi$. If $\phi$ is satisfiable in an evidential world over
$\Phih$ and $\Phio$, then $\phi$ is satisfiable in an evidential
world over $\Phih'$ and $\Phio'$, where $\abs{\Phih'}=\abs{\cH}+1$ and
$\abs{\Phio'}=\abs{\cO}+1$.
\elem
\prf 
We do this in two steps, to clarify the presentation. First, we show
that we can add a single hypothesis and observation to $\Phih$ and
$\Phio$ and preserve satisfiability of $\phi$. This means that the
second step below can assume that $\Phih\ne\cH$ and
$\Phio\ne\cO$. Assume that $\phi$ is satisfied in an evidential
world $w=(h,\ob,\mu,\cE)$ over $\Phih$ and $\Phio$, so that there
exists $v$ such that $(w,v) \sat\phi$.  Let
$\Phih'=\Phih\cup\{h^*\}$, where $h^*$ is a new hypothesis not in
$\Phih$, and let $\Phio'=\Phio\cup\{\ob^*\}$, where $\ob^*$ is a new
observation not in $\Phio$. Define the evidential world
$w'=(h,\ob,\mu',\cE')$ over $\Phih'$ and $\Phio'$, where $\cE'$ and
$\mu'$ are defined as follows.  
Define the probability measure $\mu'$ by taking: 
\[ \mu'(h)=\begin{cases}
    \mu(h) & \text{if $h\in\Phih$}\\
    0 & \text{if $h=h^*$}.
           \end{cases}\]
Similarly, define the evidence space $\cE'=(\Phih',\Phio',\bmu')$
derived from $\cE=(\Phih,\Phio,\bmu)$ by
taking:
\[ \mu'_h(\ob) = \begin{cases}
     \mu_h(\ob) & \text{if $h\in\Phih$ and $\ob\in\Phio$}\\
     0 & \text{if $h\in\Phih$ and $\ob=\ob^*$}\\
     0 & \text{if $h=h^*$ and $\ob\in\Phio$}\\
     1 & \text{if $h=h^*$ and $\ob\in\ob^*$}.
                 \end{cases}\]
Thus, $\mu'_h$ extends the existing $\mu_h$ by assigning a probability
of $0$ to the new observation $\ob^*$; in contrast, the new probability
$\mu'_{h^*}$ assigns probability 1 to the new observation $\ob^*$. 
We can check that $(w',v)\sat\phi$.

The second step is to ``collapse'' all the hypotheses and observations
that do not appear in $\phi$ into one of the hypotheses that do not
appear in $\cH$ and $\cO$, which by the previous step are guaranteed
to exist. By the previous step, we can assume that $\Phih\ne\cH$ and
$\Phio\ne\cO$. Assume $\phi$ is satisfiable in an evidential world
$w=(h,\ob,\mu,\cE)$ over $\Phih$ and $\Phio$, that is, there exists
$v$ such that $(w,v)\sat\phi$. Pick an hypothesis and
an observation from $\Phih$ and $\Phio$ as follows, depending on the
hypothesis $h$ and observation $\ob$ in $w$. Let $h^\dagger$ be $h$
if $h\not\in\cH$, otherwise, let $h^\dagger$ be an arbitrary element
of $\Phih-\cH$; let $\Phih'=\cH\cup\{h^\dagger\}$. Similarly, let
$\ob^\dagger$ be $\ob$ if $\ob\not\in\cO$, otherwise, let
$\ob^\dagger$ be an arbitrary element of $\Phio-\cO$; let
$\Phio'=\cO\cup\{\ob^\dagger\}$. Let
$w'=(h,\ob,\mu',\cE')$ be an evidential world over $\Phih'$ and
$\Phio'$ obtained from $w$ as 
follows. Define the probability measure $\mu'$ by taking:
\[ \mu'(h)=\begin{cases}
     \mu(h) & \text{if $h\in\cH$}\\
     \sum_{h'\in\Phih-\cH}\mu(h') & \text{if $h=h^\dagger$}.
           \end{cases}\]
Define $\cE'=(\Phih',\Phio',\bmu')$ derived from 
$\cE=(\Phih,\Phio,\bmu)$ by taking:
\[ \mu'_h(\ob) = \begin{cases}
     \mu_h(\ob) & \text{if $h\in\cH$ and $\ob\in\cO$}\\
     \sum_{\ob'\in\Phio-\cO}\mu_h(\ob') & \text{if $h\in\cH$ and
$\ob=\ob^\dagger$}\\
     \sum_{h'\in\Phih-\cH}\mu_{h'}(\ob) & \text{if $h=h^\dagger$ and
$\ob\in\cO$}\\
     \sum_{h'\in\Phih-\cH}\sum_{\ob'\in\Phio-\cO}\mu_{h'}(\ob') &
\text{if $h=h^\dagger$ and $\ob=\ob^\dagger$}.
                 \end{cases}\]
We can check by induction that $(w',v)\sat\phi$. 
\eprf

\othm{t:complexity1} 
There is a procedure that runs in space exponential in
$\abs{\phi}~\norm{\phi}$ for deciding, given $\Phih$ and $\Phio$,
whether a formula $\phi$ of $\Lfoev(\Phih,\Phio)$ is satisfiable in an
evidential world.
\eothm
\prf
Let $\phi$ be a formula of $\Lfoev(\Phih,\Phio)$. 
By Lemma~\ref{l:smallsets}, $\phi$ is satisfiable if we can construct
a probability measure $\mu$ on $\Phih'=\mathcal{H}\cup\{h^*\}$ (where
$\mathcal{H}$ is the set of hypotheses appearing in $\phi$, and
$h^*\not\in\mathcal{H}$) and probability measures 
$\mu_{h_1},\dots,\mu_{h_m}$ on $\Phio'=\mathcal{O}\cup\{\ob^*\}$
(where $\mathcal{O}$ is the set of observations appearing in $\phi$
and $\ob^*\not\in\mathcal{O}$) such that
$\cE=(\Phih',\Phio',\bmu)$, $w=(h,\ob,\mu,\cE)$ with $(w,v)\sat\phi$ for
some $h$, $\ob$, and $v$.

The aim now is to derive a formula $\phi'$ in the language of real
closed fields that asserts the existence of these probability
measures. More precisely, we can adapt the construction of the formula
$\phi'$ from $\phi$ in the proof of
Theorem~\ref{t:sound-complete}. The one change we need to make is
ensure that $\phi'$ is polynomial in the size of $\phi$, which the
construction in the proof of Theorem~\ref{t:sound-complete} does not
guarantee. The culprit is the fact that we encode integer constants
$k$ as $1+\dots+1$. It is straightforward to modify the construction
so that we use a more efficient representation of integer constants,
namely, a binary representation. For example, we can write $42$ as
$2(1+2^2(1+2^2))$, which can be expressed in the language of real closed
fields as $(1+1)(1+(1+1)(1+1)(1+(1+1)(1+1)))$. We can
check that if $k$ is a coefficient of length $k$ (when written in
binary), it can be written as a term of length $O(k)$ in the language of
real closed fields. 
Thus, we modify the construction of $\phi'$ in the proof of
Theorem~\ref{t:sound-complete} so that integer constants $k$ are
represented using the above binary encoding. It is easy to see that
$\abs{\phi'}$ is polynomial in $\abs{\phi}~\norm{\phi}$ 
(since $|\Phih'|$ and $|\Phio'|$ are both polynomial in $|\phi|$). 
We can now use the exponential-space algorithm of Ben-Or, Kozen, and
Reif \citeyear{BKR} on
$\phi'$: if $\phi'$ is satisfiable, then we can construct the
required probability measures, and $\phi$ is satisfiable; otherwise,
no such probability measures exist, and $\phi$ is unsatisfiable. 
\eprf

\othm{t:complexity1-alt}
There is a procedure that runs in space exponential in
$\abs{\phi}~\norm{\phi}$ for deciding whether
there exist sets of primitive propositions $\Phih$ and $\Phio$ such
that $\phi\in\Lfoev(\Phih,\Phio)$ and $\phi$ is satisfiable in an
evidential world.
\eothm
\prf
Let $h_1,\dots,h_m$ be the hypotheses appearing in $\phi$, and
$\ob_1,\dots,\ob_n$ be the hypotheses appearing in $\phi$. Let
$\Phih=\{h_1,\dots,h_m,h^*\}$ and $\Phio=\{\ob_1,\dots,\ob_n,\ob^*\}$,
where $h^*$ and $\ob^*$ are an hypothesis and observation not
appearing in $\phi$. Clearly, $\abs{\Phih}$ and $\abs{\Phio}$ are
polynomial in $\abs{\phi}$. By Lemma~\ref{l:smallsets}, if $\phi$ is
satisfiable in an evidential world, it is satisfiable in an
evidential world over $\Phih$ and $\Phio$. By
Theorem~\ref{t:complexity1}, we have an algorithm
to determine if $\phi$ is satisfied in an evidential world over
$\Phih$ and $\Phio$
that runs in  space exponential in $\abs{\phi}~\norm{\phi}$. 
\eprf

\othm{t:complexity2} 
There is a procedure that runs in space
polynomial in $\abs{\phi}~\norm{\phi}$ for deciding, given $\Phih$ and
$\Phio$, whether a formula $\phi$ of $\Lev(\Phih,\Phio)$ is
satisfiable in an evidential world.  
\eothm
\prf
The proof of this result is very similar to that of
Theorem~\ref{t:complexity1}. 
Let $\phi$ be a formula of $\Lev(\Phih,\Phio)$. 
By Lemma~\ref{l:smallsets}, $\phi$ is satisfiable if 
there exists
a probability measure $\mu$ on $\Phih'=\mathcal{H}\cup\{h^*\}$ (where
$\mathcal{H}$ is the set of hypotheses appearing in $\phi$, and
$h^*\not\in\mathcal{H}$), probability measures 
$\mu_{h_1},\dots,\mu_{h_m}$ on $\Phio'=\mathcal{O}\cup\{\ob^*\}$
(where $\mathcal{O}$ is the set of observations appearing in $\phi$
and $\ob^*\not\in\mathcal{O}$),
a hypothesis $h$, observation $o$, and valuation $v$ such that
$(w,v)\sat\phi$, where $w=(h,\ob,\mu,\cE)$ and $\cE=(\Phih',\Phio',\bmu)$. 

We derive a formula $\phi'$ in the language of real
closed fields that asserts the existence of these probability
measures by adapting the construction of the formula $\phi'$ from
$\phi$ in the proof of Theorem~\ref{t:sound-complete}. 
As in the proof of Theorem~\ref{t:complexity1}, we need to make sure
that $\phi'$ is polynomial in the size of $\phi$, which the
construction in the proof of Theorem~\ref{t:sound-complete} does not
guarantee. 
We modify the construction so that we use a more
efficient representation of integer constants, namely, a binary
representation. For example, we can write $42$ as $2(1+2^2(1+2^2))$,
which can be expressed in the language of real closed fields as
$(1+1)(1+(1+1)(1+1)(1+(1+1)(1+1)))$. We can check that if $k$ is a
coefficient of length $k$ (when written in binary), it can be written
as a term of length $O(k)$ in the language of real closed fields. 
We modify the construction of $\phi'$ in the proof of
Theorem~\ref{t:sound-complete} so that integer constants $k$ are
represented using this binary encoding. It is easy to see that
$\abs{\phi'}$ is polynomial in $\abs{\phi}~\norm{\phi}$ 
(since $|\Phih'|$ and $|\Phio'|$ are both polynomial in $|\phi|$). 
The key now is to notice that the resulting formula $\phi'$ can be
written as $\exists x_1\dots \exists x_n(\phi'')$ for some quantifier-free
formula $\phi''$. 
In this form, we can apply the polynomial space algorithm of Canny
\citeyear{r:canny88} to $\phi''$: if $\phi''$ is satisfiable, then we can
construct the required probability measures, and $\phi$ is
satisfiable; otherwise, no such probability measures exist, and $\phi$
is unsatisfiable. 
\eprf

\othm{t:complexity2-alt}
There is a procedure that runs in space polynomial in
$\abs{\phi}~\norm{\phi}$ for deciding whether there exists sets of
primitive  propositions $\Phih$ and $\Phio$ such that 
$\phi\in\Lev(\Phih,\Phio)$ and $\phi$ is satisfiable in an
evidential world.
\eothm
\prf
Let $h_1,\dots,h_m$ be the hypotheses appearing in $\phi$, and
$\ob_1,\dots,\ob_n$ be the hypotheses appearing in $\phi$. Let
$\Phih=\{h_1,\dots,h_m,h^*\}$ and $\Phio=\{\ob_1,\dots,\ob_n,\ob^*\}$,
where $h^*$ and $\ob^*$ are an hypothesis and observation not
appearing in $\phi$. 
Clearly, $\abs{\Phih}$ and $\abs{\Phio}$ are polynomial in
$\abs{\phi}$. 
By Lemma~\ref{l:smallsets}, if $\phi$ is satisfiable in an evidential
world, it is satisfiable in an evidential world over $\Phih$ and
$\Phio$. 
By Theorem~\ref{t:complexity2}, we have an algorithm to determine if
$\phi$ is satisfied in an evidential world over $\Phih$ and $\Phio$
that runs in space polynomial in $\abs{\phi}~\norm{\phi}$.  
\eprf

The proofs of Theorem~\ref{t:complexity3} and \ref{t:complexity3-alt}
rely on the following small model result, a variation on
Lemma~\ref{l:smallsets}. 

\lem\label{l:smallworld}
Given $\Phih$ and $\Phio$, let $\phi$ be a formula of
$\Lfoev(\Phih,\Phio)$, and let $\cH\subseteq\Phih$ and
$\cO\subseteq\Phio$ be the hypotheses and observations that occur in
$\phi$. If $\phi$ is satisfiable in an evidential world over
$\Phih$ and $\Phio$, then $\phi$ is satisfiable in an evidential
world over $\Phih'$ and $\Phio'$ where $\abs{\Phih'}=\abs{\cH}+1$ and
$\abs{\Phio'}=\abs{\cO}+1$, and where, for each $h\in\Phih'$ and 
$\ob\in\Phio'$, the likelihood $\mu_h(\ob)$ is a rational number with
size $O(\abs{\phi}~\norm{\phi}+\abs{\phi}\log(\abs{\phi}))$. 
\elem
\prf
Let $\phi$ be a formula satisfiable in an evidential world over
$\Phih$ and $\Phio$. 
By Lemma~\ref{l:smallsets}, $\phi$ is satisfiable in an evidential
world over $\Phih'$ and $\Phio'$, where $\abs{\Phih'}=\abs{\cH}+1$ and
$\abs{\Phio'}=\abs{\cO}+1$. 
To force the likelihoods to be small, we adapt Theorem~2.6 in
FHM, which says that if a formula $f$ in the FHM logic is satisfiable, it
is satisfiable in a structure where the probability assigned to each
state of the structure is a rational number with size
$O(\abs{f}~\norm{f}+\abs{f}\log(\abs{f}))$. The formulas in
$\Lw(\Phih',\Phio')$ are just formulas in the FHM logic.
The result adapts immediately, and yields the required bounds for 
the size of 
the likelihoods.
\eprf

\othm{t:complexity3}
The problem of deciding, given $\Phih$ and $\Phio$,  whether a formula
$\phi$ of $\Lw(\Phih,\Phio)$ is satisfiable in an evidential world
is NP-complete. 
\eothm
\prf
To establish the lower bound, 
observe that we can reduce propositional satisfiability to
satisfiability in $\Lw(\Phih,\Phio)$. 
More precisely, let $f$ be a propositional formula, where
$p_1,\dots,p_n$ are the primitive propositions appearing in $f$. 
Let $\Phio=\{\ob_1,\dots,\ob_n,\ob^*\}$ be a set of observations, where
observation $\ob_i$ corresponds to the primitive proposition $p_i$, and
$\ob^*$ is another (distinct) observation; let $\Phih$ be an arbitrary
set of hypotheses, and let $h$ be an arbitrary hypothesis in $\Phih$.
Consider the formula $\hat{f}$ obtained by replacing every occurrence
of $p_i$ in $f$ by $\We(\ob_i,h)>0$. 
It is straightforward to verify that $f$ is satisfiable if and only if
$\hat{f}$ is satisfiable in $\Lw(\Phih,\Phio)$. 
(We need the extra observation $\ob^*$ to take care of the case
$f$ is satisfiable in a a model where each of $p_1, \ldots, p_n$ is
false.   In that case, $\We(\ob_1,h) = \cdots \We(\ob_n,h) = 0$, but we can
take $\We(\ob^*,h) = 1$.)  This establishes the lower bound, 

The upper bound is straightforward. By Lemma~\ref{l:smallworld}, an
evidential world over $\Phih$ and $\Phio$ can be guessed in time
polynomial in $\abs{\Phih}+\abs{\Phio}+\abs{\phi}~\norm{\phi}$, since
the prior probability in the world requires assigning a value to
$\abs{\Phih}$ hypotheses, and the evidence space requires
$\abs{\Phih}$ likelihood functions, each assigning a value to
$\abs{\Phio}$ observations, of size polynomial in
$\abs{\phi}~\norm{\phi}$. 
We can verify that a world satisfies $\phi$ in time polynomial in 
$\abs{\phi}~\norm{\phi}+\abs{\Phih}+\abs{\Phih}$. This establishes
that the problem is in NP.
\eprf

\othm{t:complexity3-alt}
The problem of deciding, for a formula $\phi$, whether there exists
sets of primitive propositions $\Phih$ and $\Phio$ such that
$\phi\in\Lw(\Phih,\Phio)$ and $\phi$ is satisfiable in an
evidential world is NP-complete.
\eothm
\prf
For the lower bound, we reduce from the decision problem of
$\Lw(\Phih,\Phio)$ over fixed $\Phih$ and $\Phio$. Let
$\Phih=\{h_1,\dots,h_m\}$ and $\Phio=\{\ob_1,\dots,\ob_n\}$, and let
$\phi$ be a formula in $\Lw(\Phih,\Phio)$. We can check that $\phi$
is satisfiable in evidential world over $\Phih$ and $\Phio$ if and 
only if $\phi\land(h_1\lor\dots\lor h_m)\land(\ob_1\lor\dots\lor \ob_n)$ 
is satisfiable in an evidential world over arbitrary $\Phih'$ and
$\Phio'$. Thus, by Theorem~\ref{t:complexity3}, we get our lower
bound. 

For the upper bound, by Lemma~\ref{l:smallworld}, if $\phi$ is
satisfiable, it is satisfiable in an evidential world over $\Phih$ and
$\Phio$, where $\Phih=\cH\cup\{h^*\}$,  $\cH$ consists of the
hypotheses appearing in $\phi$, $\Phio=\cO\cup\{\ob^*\}$, $\cO$
consists of the observations appearing in $\phi$, and $h^*$ and
$\ob^*$ are new hypotheses and observations. 
Thus, $\abs{\Phih}\le\abs{\phi}+1$, and $\abs{\Phio}\le\abs{\phi}+1$. 
As in the proof of Theorem~\ref{t:complexity3}, such a world can be
guessed in time polynomial in
$\abs{\phi}~\norm{\phi}+\abs{\Phih}+\abs{\Phio}$, and therefore in time
polynomial in $\abs{\phi}~\norm{\phi}$. We can verify that this world
satisfies $\phi$ in time polynomial in $\abs{\phi}~\norm{\phi}$,
establishing that the problem is in NP.
\eprf

\othm{t:sound-complete-dyn}
$\AXdyn$ is a sound and complete axiomatization for
$\Lfoevdyn(\Phih,\Phio)$ with respect to evidential runs.
\eothm
\prf
It is easy to see that each axiom is valid in evidential
runs. To prove completeness, we follow the same procedure as in the 
proof of Theorem~\ref{t:sound-complete}, showing that if $\phi$ is
consistent, then it is satisfiable, that is, there exists an
evidential run $r$ and valuation $v$ such that
$(r,m,v)\sat\phi$ for some point $(r,m)$ of $r$.

As in the body of the paper, let $\Phih=\{h_1,\dots,h_{\nh}\}$ and
$\Phio=\{\ob_1,\dots,\ob_{\no}\}$. Let $\phi$ be a consistent formula. 
The first step of the process is to reduce the formula $\phi$ to a
canonical form with respect to the $\Circ$ operator. Intuitively, we
push down every occurrence of a $\Circ$ to the polynomial
inequality formulas present in the formula. It is easy to see that
axioms 
and inference rules
\axiom{T1}--\axiom{T6} can be used to establish that $\phi$ is provably equivalent
to a formula $\phi'$ where every occurrence of $\Circ$ is in the form
of subformulas $\Circ^n(\ob)$ and $\Circ^n(p\ge c)$, where $p$ is a
polynomial term that contains at least one occurrence of the $\Pr$
operator. We use the notation $\Circ^n\phi$ for
$\Circ\dots\Circ\phi$, the $n$-fold application of $\Circ$ to $\phi$.
We write $\Circ^0\phi$ for $\phi$. Let $N$ be the maximum coefficient
of $\Circ$ in $\phi'$.  

By way of contradiction, assume that
$\phi'$ (and hence $\phi$) is unsatisfiable. 
As in the proof of Theorem~\ref{t:sound-complete}, 
we reduce the formula $\phi'$ to an
equivalent formula in the language of real closed fields.  Let
$u_1,\dots,u_{\nh}$, $v^0_1,\dots,v^0_{\no},\dots,v^N_1,\dots,v^N_{\no}$,
$y^0_1,\dots,y^0_{\no},\dots,y^N_1,\dots,y^N_{\no}$, and
$z_{\<i_1,\dots,i_k\>,1},\dots,z_{\<i_1,\dots,i_k\>,\nh}$ (for every
sequence $\<i_1,\dots,i_k\>$) be new variables, where, intuitively,
\begin{itemize}
\item $u_i$ gets value $1$ if hypothesis $h_i$ holds, $0$ otherwise;
\item $v^n_i$ gets value $1$ if observation $\ob_i$ holds at time $n$, $0$
otherwise;
\item $y^n_i$ represents $\Pr(h_i)$ at time $n$;
\item $z_{\<i_1,\dots,i_k\>,j}$ represents $\We(\<\ob^{i_1},\dots,\ob^{i_k}\>,h_j)$.
\end{itemize}
The main difference with the construction in the proof of
Theorem~\ref{t:sound-complete} is that we have variables $v_i^n$
representing the observations at every time step $n$, rather than
variables representing observations at the only time step, variables
$y_i^n$ representing each hypothesis probability at every time step,
rather than variables representing prior and posterior probabilities,
and variables $z_{\<i_1,\dots,i_k\>,j}$ representing the weight of
evidence of sequences of observations, rather than variables
representing the weight of evidence of single observations.  Let
$\mathbf{v}$ represent that list of new variables. We consider the
same formulas as in the proof of Theorem~\ref{t:sound-complete},
modified to account for the new variables, and the fact that we are
reasoning over multiple time steps. More specifically, the formula
$\phihyp$ is unchanged. Instead of $\phiobs$, we consider formulas
$\phiobs^1,\dots,\phiobs^N$ saying that exactly one observation holds
at each time time step, where $\phiobs^n$ is given by:
\[ (v^n_1=0\lor v^n_1=1)\land\dots\land(v^n_{\no}=0\lor
v^n_{\nh}=1)\land v^n_1+\dots+v^n_{\nh}=1.\]
Let $\phiobs'=\phiobs^1\land\dots\land\phiobs^N$.

Similarly, instead of $\phiprior$ and $\phipost$, we consider formulas
$\phiprob^1,\dots,\phiprob^N$ expressing that $\Pr$ is
a probability measure at each time step, where $\phiprob^n$ is given by:
\[ y^n_1\ge 0\land\dots\land y^n_{\nh}\ge 0\land
y^n_1+\dots+y^n_{\nh}=1.\]
Let $\phiprob=\phiprob^1\land\dots\land\phiprob^N$.

Similarly, we consider $\phiwprob$ and $\phiwf$, except where we
replace variables $z_{i,j}$ by $z_{\<i\>,j}$, to reflect the fact that
we now consider sequences of observations. The formula $\phiwup$,
capturing the update of a prior probability into a posterior
probability
given by \axiom{E5}, 
is replaced by the formulas $\phiwup^1,\dots,\phiwup^N$
representing the update of the probability at each time step, where
$\phiwup^n$ is given by the obvious generalization of $\phiwup$:
\begin{align*}
  & (v^n_1=1 \rimp (\begin{prog}
                 y^{n-1}_1 z_{1,1} = y^n_1 y^{n-1}_1 z_{1,1} + \dots +
                 y^n_1 y^{n-1}_{\nh} z_{1,\nh}\, \land\\ 
                 \dots\land
                 y^{n-1}_{\nh} z_{1,\nh} = y^n_{\nh} y^{n-1}_1 z_{1,1} + \dots + y^n_{\nh} y^{n-1}_{\nh} z_{1,\nh}))\land
                 \end{prog}\\
  & \dots\land\\
  & (v^n_{\no}=1 \rimp (\begin{prog}
                     y^{n-1}_1 z_{\no,1} = y^n_1 y^{n-1}_1 z_{\no,1} +
                 \dots + y^n_1 y^{n-1}_{\nh} z_{\no,\nh}\, \land\\ 
                     \dots\land
                     y^{n-1}_{\nh} z_{\no,\nh} = y^n_{\nh}
y^{n-1}_1 z_{\no,1} + \dots y^n_{\nh} y^{n-1}_{\nh} z_{\no,\nh})).
           \end{prog}
\end{align*}
Let $\phiwup'=\phiwup^1\land\dots\land\phiwup^N$.

Finally, we need a new formula $\phiwcomp$ capturing the relationship between the
weight of evidence of a sequence of observations, and the weight of
evidence of the individual observations, to capture axiom~\axiom{E6}: 
\begin{align*}
   & \bigwedge\limits_{\substack{1\le k \le N\\ 1\le
i_1,\dots,i_k\le\no}} 
     z_{\<i_1\>,h_1}\cdots z_{\<i_k\>,h_1}=\begin{prog}
        z_{\<i_1,\dots,i_k\>,h_1}z_{\<i_1\>,h_1}\cdots
z_{\<i_k\>,h_1}\\ +\dots 
+z_{\<i_1,\dots,i_k\>,h_1}z_{\<i_1\>,h_{\nh}}\cdots
z_{\<i_k\>,h_{\nh}}\land\end{prog}\\ 
   & \dots\land
 \bigwedge\limits_{\substack{1\le k \le N\\ 1\le
i_1,\dots,i_k\le\no}} z_{\<i_1\>,h_{\nh}}\cdots z_{\<i_k\>,h_{\nh}}=
\begin{prog}
    z_{\<i_1,\dots,i_k\>,h_{\nh}}z_{\<i_1\>,h_1}\cdots
z_{\<i_k\>,h_1}\\ +\dots
+z_{\<i_1,\dots,i_k\>,h_{\nh}}z_{\<i_1\>,h_{\nh}}\cdots
z_{\<i_k\>,h_{\nh}}.
\end{prog}
\end{align*}

Let $\hat{\phi}$ be the formula in the language of real closed fields
obtained from $\phi$ by replacing each occurrence of the primitive
proposition $h_i$ by $u_i=1$, each occurrence of $\Circ^n\ob_i$ by
$v^n_i=1$, and within each polynomial inequality formula $\Circ^n(p\ge 
c)$, replacing each occurrence of $\Pr(\rho)$ by $\sum_{h_i\in\intension{\rho}}y^n_i$,
each occurrence of $\We(\<\ob^{i_1},\dots,\ob^{i_k}\>,h_j)$ by
$z_{\<i_1,\dots,i_k\>,j}$, 
and each occurrence of an integer
coefficient $k$ by $1+\dots+1$ ($k$ times).  
Finally, let $\phi'$ be the formula
$\exists\mathbf{v}(\phihyp\land\phiobs'\land\phiprob\land\phiwprob\land\phiwf\land\phiwup'\land\phiwcomp\land\hat{\phi})$.

It is easy to see that if $\phi$ is unsatisfiable over evidential
systems,  then $\phi'$ is false about the real numbers. Therefore,
$\neg\phi'$ must be a formula valid in real closed fields, and
hence an instance of \axiom{RCF}. Thus, $\neg\phi'$ is provable. It is
straightforward to show, using the obvious variant of
Lemma~\ref{l:decompose} 
that $\neg\phi$ itself is provable, contradicting the fact that $\phi$
is consistent. Thus, $\phi$ must be satisfiable, establishing
completeness.
\eprf

\bibliographystyle{theapa}
\bibliography{riccardo2,z,refs,joe,bghk}

\end{document}